\documentclass[10pt,twocolumn,letterpaper]{article}

\usepackage{cvpr}
\usepackage{times}
\usepackage{epsfig}
\usepackage{graphicx}
\usepackage{amsmath}
\usepackage{amssymb}
\usepackage{algorithm}  
\usepackage{algpseudocode} 
\usepackage{caption}
\usepackage{subfigure}
\usepackage{multirow}

\usepackage[pagebackref=true,breaklinks=true,letterpaper=true,colorlinks,bookmarks=false]{hyperref}

\cvprfinalcopy % *** Uncomment this line for the final submission

\def\cvprPaperID{2846} % *** Enter the CVPR Paper ID here

\ifcvprfinal\fi
\begin{document}

%%%%%%%%% TITLE
\title{Recurrent Feature Reasoning for Image Inpainting}

\author{Jingyuan Li${}^1$, Ning Wang${}^1$, Lefei Zhang*${}^1$, Bo Du${}^1$, Dacheng Tao${}^2$\\
\\
${}^1$ National Engineering Research Center for Multimedia Software, School of Computer Science\\
and Institute of Artificial Intelligence, Wuhan University, China\\
${}^2$ UBTECH Sydney AI Centre, School of Computer Science, Faculty of Engineering,\\ The University of Sydney, Darlington, NSW 2008, Australia\\
{\tt \small  $\{$jingyuanli, wang\_ning, zhanglefei, dubo$\}$@whu.edu.cn, dacheng.tao@sydney.edu.au}
%\and
%Dacheng Tao\\
%{\tt  secondauthor@i2.org}
}

\twocolumn[{%
\renewcommand\twocolumn[1][]{#1}%
\maketitle
%\begin{center}
\centering
\includegraphics[scale=0.35]{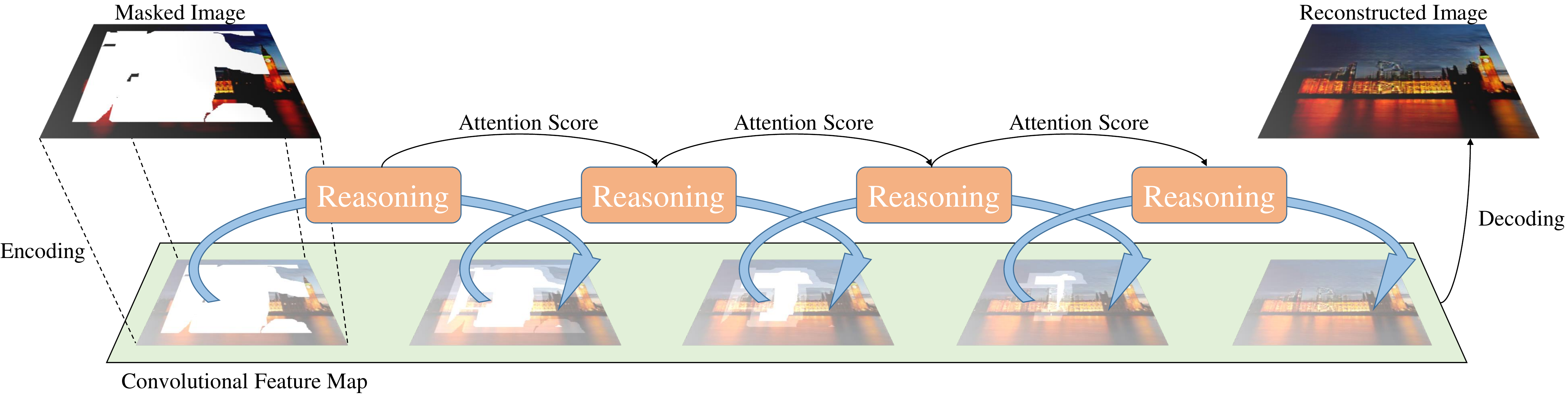}
\label{rfrinp}
\captionof{figure}{The overview of our proposed inpainting scheme. The masked image is first mapped into the convolutional feature space and processed by a shared Feature Reasoning module recurrently. After the feature map is fully recovered, the generated feature maps are merged together (Omitted in this figure) and the merged feature is translated back to a RGB image.}
%\end{center}
\vspace{0.33cm}
}]
\thispagestyle{empty}
\renewcommand{\thefootnote}{\fnsymbol{footnote}}
\footnotetext[1]{Corresponding Author}
%%%%%%%%% ABSTRACT
\begin{abstract}
	Existing inpainting methods have achieved promising performance for recovering regular or small image defects. However, filling in large continuous holes remains difficult due to the lack of constraints for the hole center. In this paper, we devise a \textbf{Recurrent Feature Reasoning} (RFR) network which is mainly constructed by a plug-and-play Recurrent Feature Reasoning module and a \textbf{Knowledge Consistent Attention} (KCA) module. Analogous to how humans solve puzzles (i.e., first solve the easier parts and then use the results as additional information to solve difficult parts), the RFR module recurrently infers the hole boundaries of the convolutional feature maps and then uses them as clues for further inference. The module progressively strengthens the constraints for the hole center and the results become explicit. To capture information from distant places in the feature map for RFR, we further develop KCA and incorporate it in RFR. Empirically, we first compare the proposed RFR-Net with existing backbones, demonstrating that RFR-Net is more efficient (e.g., a 4\% SSIM improvement for the same model size). We then place the network in the context of the current state-of-the-art, where it exhibits improved performance. The corresponding source code is available at: \url{https://github.com/jingyuanli001/RFR-Inpainting}
\end{abstract}

%%%%%%%%% BODY TEXT
\vspace{-0.5cm}
\section{Introduction}
Image inpainting aims to recover the missing regions of damaged images with realistic contents. These algorithms have a wide range of applications in photo editing, de-captioning and other scenarios where people might want to remove unwanted objects from their photos \cite{song2018,shetty2018,barnes2009,criminisi2004}. A successfully inpainted image should exhibit coherence in both structure and texture between the estimated pixels and the background \cite{yan2018, nazeri2019}.

\begin{figure}[t]
	\centering
	\includegraphics[scale=0.38]{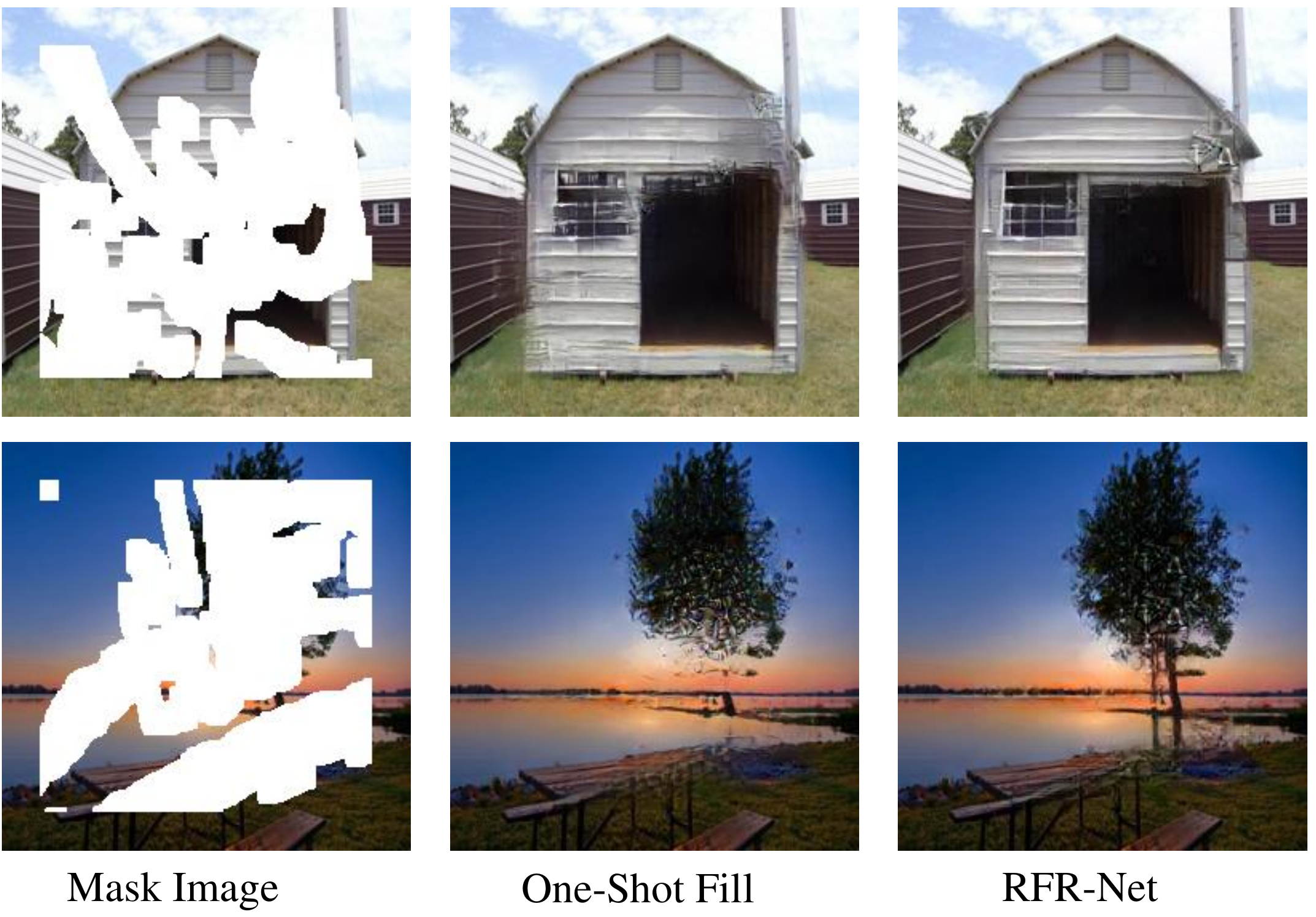}
	\vspace{-8pt}
	\caption{Illustration of semantic ambiguity that existing one-shot inpainting methods suffer from. Due to the lack of explicit constraints on the hole center, it is challenging for a network to directly inpaint on the hole center. The results of One-Shot Fill are taken from the state-of-the-art method \cite{li2019}.}
	\vspace{-16pt}
	\label{figure2}
\end{figure}

Recently, deep convolutional networks have been used to solve the inpainting problem. Most state-of-the-art (SOTA) methods~\cite{yu2018,nazeri2019,xie2019,yu2019} exploit an encoder-decoder architecture and assume that the damaged image, once encoded, should have adequate information for reconstruction and then inpaint in one shot. This assumption is reasonable for small or narrow defects, because pixels within local areas can have strong correlations and a pixel can therefore be inferred from its surroundings. However, as the damaged areas become larger and the distances between known and unknown pixels increase, these correlations are weakened and the constraints for the hole center loosen. Under such circumstances, the information in the known area is not informative for recovering the pixels at the center of the hole and the networks generate semantically ambiguous results (see Fig \ref{figure2}). An alternative scheme is to inpaint in a progressive fashion from the hole boundary to the center \cite{zhang2018,guo2019}. However, these methods do not use recurrent designs and render redundant models. Also, because progressive inpainting is performed at the image level, the computational cost makes these methods less practical. Further, their inpainting schemes are only feasible in one-stage methods but unsuitable for multi-stage methods whose sub-networks cannot meet the design requirement that inputs and outputs are represented in the same space (e.g., RGB space). Finally, the process of mapping the feature map back to the RGB space occurs in each iteration, which results in information distortion in every recurrence (e.g., transforming a 128$\times$128$\times$64 feature map into a 256$\times$256$\times$3 RGB image). As a result, they either underperform or have an unacceptably high computational cost.

In this paper, we propose a new deep image inpainting architecture called the Recurrent Feature Reasoning Network (RFR-Net). Specifically, we devise a plug-and-play Recurrent Feature Reasoning (RFR) module to recurrently infer and gather the hole boundary for the encoded feature map. In this way, the constraints determining the internal contents are progressively strengthened and the model can produce semantically explicit results. Unlike existing progressive methods, the RFR module performs this progressive process in the feature map space, which not only ensures superior performance but also addresses the limitation that the inputs and outputs of the network need to be represented in the same space. The recurrent design reuses the parameters to deliver a much lighter model. Also, the computational cost can be flexibly controlled by moving the module up and down in the network. This is essential for building high-resolution inpainting networks, because avoiding the computation of the first and last few layers can eliminate most of the computational burden.

To further reinforce RFR's potential for recovering high-quality feature maps, attention modules~\cite{yu2018} are necessary. However, directly applying existing attention designs in the RFR is suboptimal, because they fail to consider the consistency requirements between feature maps in different recurrences. This could lead to blurred texture in recovered area. To overcome this problem, we devise a Knowledge Consistent Attention (KCA) mechanism, which shares the attention scores between recurrences and adaptively combines them to guide a patch-swap process. Assisted by the KCA, the level of consistency is enhanced and the model's performance improves.

Our main contributions are as follows:
\begin{itemize}
\item We propose a Recurrent Feature Reasoning (RFR) module, which exploits the correlation between adjacent pixels and strengthens the constraints for estimating deeper pixels. The RFR module not only significantly enhances network performance but also bypasses several limitations of progressive methods.
\item We develop a Knowledge Consistent Attention (KCA) module which adaptively combines the scores from different recurrences and ensures consistency between patch-swapping processes among recurrences, leading to better results with exquisite details.
\item The new modules are assembled and form a novel RFR network. We analyze our model in terms of both efficiency and performance and demonstrate RFR's superiority to several state-of-the-art methods in benchmark datasets.
\end{itemize}

\begin{figure*}[t]
	\centering
	\includegraphics[scale=0.33]{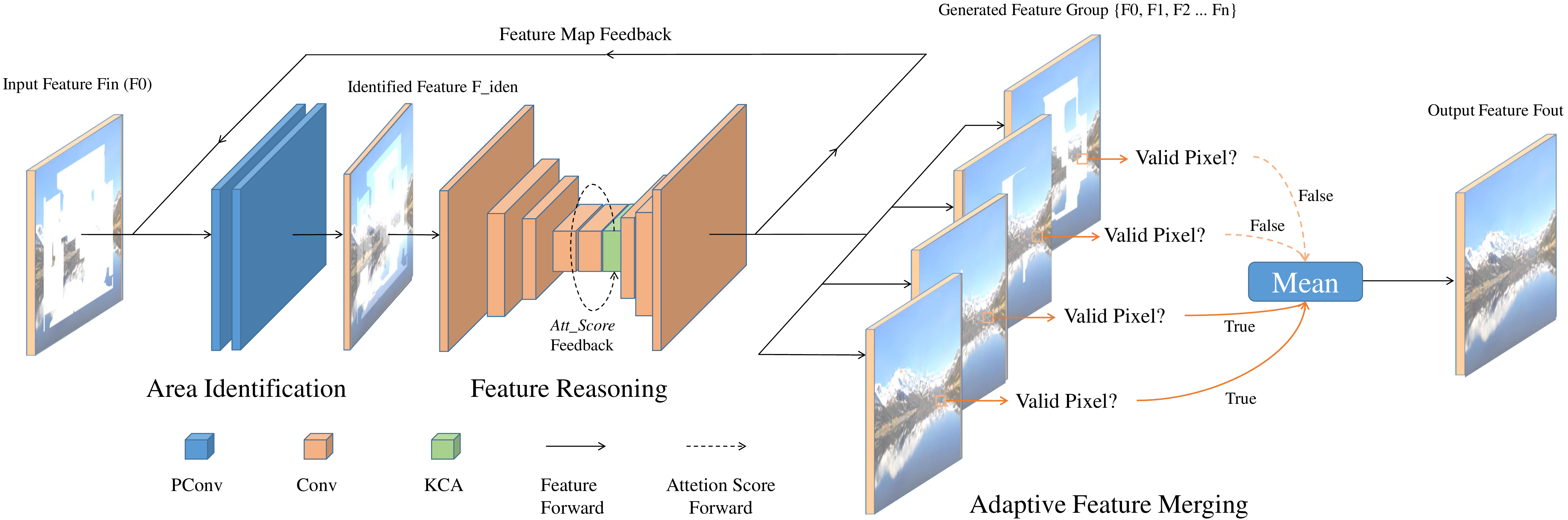}
	\caption{Illustration of the Recurrent Feature Reasoning module. The area identification process and the feature reasoning process are performed continuously. After several times of reasoning, the feature maps are merged in an adaptive fashion and an output feature map of a fixed channel numbers are generated. The module is Plug-In-and-Play and can be placed in any layer of an existing network.}
	\vspace{-10pt}
	\label{figure3}
\end{figure*}

\section{Related Works}
In this section, we summarize some previous work related to our method.

\noindent\textbf{Image Inpainting}
Traditionally, inpainting algorithms attempt to find patches from the background area to inpaint the hole ~\cite{barnes2009, bertalmio2003,chan2001,ding2019,li2016,song2018,fang2019}. Although these traditional methods are very good for simple cases, they cannot inpaint on complex scenes due to a lack of semantic understanding of the image. 

To semantically inpaint damaged images, deep convolutional networks\cite{yeh2017}, especially generative adversarial networks (GANs) \cite{goodfellow2014}, have recently been used. Context-Encoder~\cite{pathak2016} first employed a conditional GAN \cite{mirza2014} for image inpainting and demonstrated the potential of CNNs for inpainting tasks. Iizuka \etal \cite{iizuka2017} introduced an extra discriminator to ensure local image coherency and used Poisson blending \cite{perez2003} to refine the image, which rendered more detailed and sharper results. Yan \etal \cite{yan2018} and Yu \etal \cite{yu2018} devised feature shift and contextual attention operations, respectively, to allow the model to borrow feature patches from distant areas of the image. Liu \etal~\cite{liu2018} and Yu \etal \cite{yu2019} devised special convolutional layers to enable the network to inpaint on irregularly masked images. These methods fail to address the semantic ambiguity as they try to recover the whole target with inadequate constraints.

\noindent\textbf{Progressive Inpainting}
Progressive image inpainting has recently been investigated. Xiong \etal \cite{xiong2019} and Nazeri \etal \cite{nazeri2019} filled images with contour/edge completion and image completion in a step-wise manner to ensure structural consistency. Li \etal \cite{li2019} added progressively reconstructed boundary maps as extra training targets to assist the inpainting process of a U-Net. These approaches attempted to solve inpainting tasks by adding structural constraints, but they still suffer from the lack of information for restoring deeper pixels in holes for their backbone. Zhang \etal \cite{zhang2018} used cascaded generators to progressively fill in the image. Guo \etal \cite{guo2019} directly inpainted on the original size images with a single stage feed-forward network. Oh \etal \cite{oh2019} used an onion-peel scheme to progressively inpaint on video data using content from reference frames, allowing accurate content borrowing. However, these methods generally suffered from the limitations of progressive inpainting described in the introduction.

\noindent\textbf{Attentive Inpainting}
Image inpainting models can adopt an attention mechanism to borrow features from the background. Yu \etal \cite{yu2018} exploited textural similarities within the same image to fill defects with more realistic textures from the background area. Wang \etal \cite{wang2019} devised a multi-scale attention module to enhance the precision of the patch-swapping process. Liu \etal \cite{liu2019} used a coherent semantic attention layer to ensure semantic relevance between swapped features. Xie \etal \cite{xie2019} devised a bidirectional attention map estimating module for feature re-normalization and mask-updating during feature generation. Although these methods have delivered considerable improvements, they remain suboptimal for a recurrent architecture as they do not take relationships between the feature maps from different recurrences into consideration.

\section{Method}
In this section, we first introduce the RFR module, which forms the main body of RFR-Net. Then we introduce the KCA scheme, which exploit the recurrent network design. Finally, the overall architecture and corresponding target functions are introduced.

\subsection{Recurrent Feature Reasoning Module}
The RFR module is a plug-and-play module with a recurrent inference design that can be installed in any part of an existing network. The RFR module can be decomposed into three parts: 1) an area identification module, which is used to identify the area to be inferred in this recurrence; 2) a feature reasoning module, which aims to infer the content in the identified area; and 3) a feature merging operator, which merges the intermediate feature maps. Inside the module, the area identification module and the feature reasoning module work alternately and recurrently. After the holes are filled, all the feature maps generated during inference are merged to produce a feature map with fixed channel numbers. We elaborate these processes below. The model pipeline of our module is shown in Fig.~\ref{figure3}.
%%%The model pipeline? The RFR pipeline? 
\subsubsection{Area Identification}
Partial convolution \cite{liu2018} is a basic module used to identify the area to be updated in each recurrence. The partial convolutional layer updates the mask and re-normalizes the feature map after the convolution calculation. More formally, the partial convolution layer can be described as follows. Let $F^*$ denote the feature map generated by the partial convolution layer. $f^*_{x,y,z}$ denotes the feature value at location $x,y$ in the $z^{th}$ channel. ${W}_{z}$ is the $z^{th}$ convolution kernel in the layer. $f_{x,y}$ and $m_{x,y}$ are the input feature patch and input mask patch (whose size is the same as the convolutional kernel) centered at location $x,y$, respectively. Then, the feature map computed by the partial convolution layer can be represented by:
\begin{equation}\footnotesize
f^*_{x,y,z}=\left\{\begin{array}{ll}
W_{z}^{T}(f_{x,y} \odot m_{x,y}\frac{sum({1})}{sum(m_{x,y})})+b, &\text{if sum(} m_{x,y}\text{) != 0}\\
0, &\textrm{else}
\end{array}\right.
\end{equation}
Similarly, the new mask value at location $i,j$ generated by the layer can be expressed as:
\begin{equation}
m^*_{x,y}=\left\{\begin{array}{ll}
1, &\text{if sum(} m_{x,y}\text{) != 0}\\
0, &\textrm{else}
\end{array}\right.
\end{equation}
Given the equations above, we are able to receive new masks whose holes are smaller after each partial convolutional layer.

For area identification in the RFR module, we cascade several partial convolutional layers together to update the mask and feature map. After passing though the partial convolutional layers, the feature maps are processed by a normalization layer and an activation function before being sent to the feature reasoning modules. We define the difference between the updated masks and the input masks as the areas to be inferred in this recurrence. The holes in the updated mask are preserved throughout this recurrence until being further shrunk in the next recurrence.
\subsubsection{Feature Reasoning}
With the area to be processed identified, the feature values in the area are estimated by the feature reasoning module. The goal of the feature reasoning module is to fill in the identified area with as high-quality feature values as possible. The high-quality features not only produce a better final result but also benefit following inferences. As a result, the feature reasoning module can be designed to be very complicated to maximize its inference capability. However, here we simply stack a few encoding and decoding layers and bridge them using skip connections so that we can intuitively show the efficiency of the feature reasoning module.

After the feature values are inferred, the feature maps will be sent to the next recurrence. Since the RFR module does not constrain the representation of the intermediate results, the updated mask and the partially inferred feature maps are directly sent to the next recurrence without further processing.

\subsubsection{Feature Merging}
\label{sec:feature-merging}
When the feature maps are completely filled (or after a specific number of recurrences), the feature maps have passed through the feature reasoning module several times. If we directly use the last feature map to generate the output, gradient vanishing can occur, and signals generated in earlier iteration are damaged. To solve this problem, we must merge the intermediate feature maps. However, using convolutional operations to do so limits the number of recurrences, because the number of channels in the concatenation is fixed. Directly summing all feature maps removes image details, because the hole regions in different feature maps are inconsistent and prominent signals are smoothed. As a result, we use an adaptive merging scheme to address the issue. The values in the output feature map are only calculated from the feature maps whose corresponding locations have been filled. Formally, let's define $F^{i}$ as the $i^{th}$ feature map generated by the feature reasoning module and $f_{x,y,z}$ as the value at location $x$,$y$,$z$ in feature map $F$. $M^{i}$ is the binary mask for feature map $F^{i}$. The value at the output feature map $\bar{F}$ can be defined as:
\begin{equation} 
\bar{f}_{x,y,z}=\frac{\sum_{i=1}^N f_{x,y,z}^{i}}{\sum_{i=1}^N m_{x,y,z}^{i}}
\end{equation}
where $N$ is the number of feature maps. In this way, feature maps of arbitrary number can be merged, which gives the RFR the potential to fill larger holes. 
\begin{figure*}[t]
	\centering
	\includegraphics[scale=0.375]{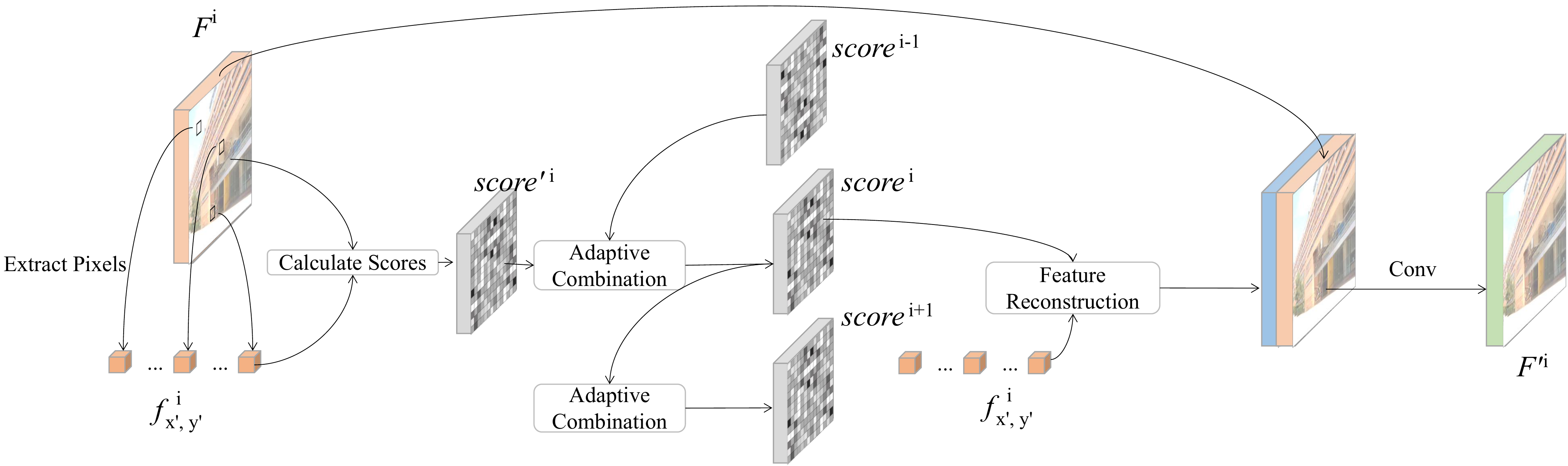}
	\caption{The illustration of the Knowledge Consistent Attention. The attention scores of the KCA are adaptively computed from the previous and current ones to ensure consistent feature ingredient.}
	\vspace{-10pt}
	\label{fig:ksa}
\end{figure*}
\subsection{Knowledge Consistent Attention}
\label{sec:ksa}
In image inpainting, the attention module is used to synthesize features of better quality \cite{yu2018}. The attention modules search for possible texture in the background and use them to replace the textures in the holes. However, directly inserting the existing attention modules into the RFR is suboptimal, because the patch swapping processes in different recurrences are performed independently. The discrepancy between the components of the synthesized feature maps can damage the feature map when they are merged (see section~\ref{sec:feature-merging}). To address the issue, we devise a novel attention module, Knowledge Consistent Attention (KCA). Unlike previous attention mechanism whose attention scores are calculated independently, the scores in our KCA are composed of scores proportionally accumulated from previous recurrences. As a result, the inconsistencies between the attention feature maps can be controlled. The illustration of the attention mechanism is given in Fig.~\ref{fig:ksa}, and the details are provided below.

In our design, the final components for each pixel are decided as follows. Let's denote the input feature map in the $i^{th}$ recurrence $F^{i}$. First, we measure the cosine similarity between each pair of feature pixels:
\begin{equation} 
\widehat{sim}_{x,y,x',y'}^{i} = \langle{\frac{f_{x,y}}{||f_{x,y}||},\frac{f_{x',y'}}{||f_{x',y'}||}}\rangle
\end{equation}
where $sim_{x,y,x',y'}^{i}$ indicates the similarity between the hole feature at location $(x, y)$ and that at location $(x', y')$. After that, we smooth the attention scores by averaging the similarity of a target pixel in an adjacent area.
\begin{equation} 
{sim'}_{x,y,x',y'}^{i} = \frac{\sum_{p,q\in\{-k,...,k\}}\widehat{sim}_{x+p,y+q,x',y'}}{k \times k}
\end{equation}
Then, we generate the proportion of the component for the pixel at location $(x, y)$ using the softmax function. The generated score map is denoted $score'$.

To calculate the final attention score of a pixel, we first make a decision about whether the score of the pixel in previous recurrence will be referred to. Given a pixel considered to be valid, its attention score in current recurrence is then calculated as the weighted sum of the original and final scores from current and previous recurrences, respectively. Formally, if the pixel at location $(x, y)$ is a valid pixel in the last recurrence (i.e., the mask value ${m}_{x, y}^{i-1}$ is 1), we adaptively combine the final score of the pixel from last recurrence with the score calculated in this recurrence as follows where $\lambda$ is a learnable parameter:
\begin{equation}
{score}_{x,y,x',y'}^{i}= \lambda {score'}_{x,y,x',y'}^{i} + (1 - \lambda) {score}_{x,y,x',y'}^{i-1}
\label{eq_weightedsum}
\end{equation}
Otherwise, if the pixel is not valid in the last recurrence, no extra operation will be performed, and the final attention score of the pixel in current recurrence will be calculated as follow:
\begin{equation} 
{score}_{x,y,x',y'}^{i}= {score'}_{x,y,x',y'}^{i}
\end{equation}
In the end, the attention score is used to rebuild the feature map. Specifically, the new feature map at location $(x, y)$ is calculated as follows:
\begin{equation} 
\widehat{f}^{i}_{x,y} = \sum_{x'\in{1, ... W} y'\in{1,...H}}score^{i}_{x,y,x',y'}f^{i}_{x',y'}
\end{equation}
After the feature map is reconstructed, the input feature $F$ and the reconstructed feature map $\widehat{F}$ are concatenated and sent to a convolution layer:
\begin{equation} 
{F'}^{i} = \phi(|\widehat{F} , F|)
\end{equation}
where $F'$ is the reconstructed feature map and $\phi$ is the pixel-wise convolution.
\subsection{Model Architecture \& Loss Functions}
\begin{algorithm}[b]
	\caption{\label{alg:rfrinp} The RFR Inpainting Network}  
	\hspace*{0.02in} {\bf Input:}\\
		${Img}_{in}$: Input image\\
		${Mask}_{in}$: Input mask\\
	\hspace*{0.02in} {\bf Output:}\\
		${Img}_{rec}$: Reconstructed image\\
	\hspace*{0.02in} {\bf Begin Algorithm}
	\begin{algorithmic}[1]
        \State $F^{0}, M^{0}$ $\gets$ $Encoding({Img}_{in}, {Mask}_{in})$ 
        \State $FeatureGroup$ $\gets$ $\{F^{0}\}$
        \State $i$ $\gets$ $0$
		\While{$i$ smaller than $IterNum$}
		\State $F^{i+1}$,$M^{i+1}$$\gets$$AreaIden(F^{i},M^{i})$
		\State $F^{i+1}$$\gets$$FeatReason(F^{i+1})$
		\State $FeatureGroup$ $\gets$ $FeatureGroup$ + $\{F^{i+1}\}$
		\State $i$$\gets$$i+1$
		\EndWhile
		\State $F_{merged}$$\gets$$FeatMerge(FeatureGroup)$
		\State ${Img}_{rec}$$\gets$$Decoding(F_{merged})$
		\State \Return ${Img}_{rec}$
	\end{algorithmic}  
\end{algorithm}
In our implementation, we put 2 and 4 convolution layers before and after the RFR module, respectively. We manually choose the recurrence number $IterNum$ to be 6 to simplify training. The KCA module is placed after the third last layer in the RFR module's feature reasoning module. The pipeline of the network is described in Alg.~\ref{alg:rfrinp}.

For image generation learning, the perceptual loss and style loss from a pre-trained and fixed VGG-16 are used. The perceptual loss and style loss compare the difference between the deep feature map of the generated image and the ground truth. Such a loss function can effectively teach the model the image's structural and textural information. These loss functions are formalized as follows. $\phi_{pool_i}$ denotes feature maps from the $i^{th}$ pooling layer in the fixed VGG-16. In the following equations, $H_{i}$, $W_{i}$ and $C_{i}$ are used to express the height, weight and channel size of the $i^{th}$ feature map, respectively. The perceptual loss can then be written as follows:
\begin{equation} 
L_{perceptual}=\sum_{i=1}^N\frac{1}{H_{i}W_{i}C_{i}}|\phi_{pool_i}^{gt}-\phi_{pool_i}^{pred}|_{1}
\end{equation}
Similarly, the computation of the style loss is as follows:
\begin{equation} 
\phi_{pool_i}^{style} = \phi_{pool_i}\phi_{pool_i}^{T}
\end{equation}
\begin{equation} 
L_{style}=\sum_{i=1}^N\frac{1}{C_{i}\times C_{i}}\big{|}\frac{1}{H_{i}W_{i}C_{i}}(\phi_{pool_i}^{style_{gt}}-\phi_{pool_i}^{style_{pred}})\big{|}_{1}
\end{equation}
Further, $L_{valid}$ and $L_{hole}$ which calculate L1 differences in the unmasked area and masked area respectively are also used in our model. In summary, our total loss function is:
\begin{equation} 
\begin{aligned}
L_{total}&=\lambda_{hole}L_{hole}+\lambda_{valid}L_{valid}+\lambda_{style}L_{style}\\
&+\lambda_{perceptual}L_{perceptual}
\end{aligned}
\end{equation}
The loss function combination in our model is similar to \cite{liu2018} and has been shown to be effective in previous works \cite{li2019}. This kind of loss function combination also enables efficient training due to the smaller number of parameters to update.

\section{Experiments}
In this section, we provide the detailed experimental settings to assist with reproducibility.

\begin{figure*}[t]
\begin{center}
  \includegraphics[width=1.0\linewidth]{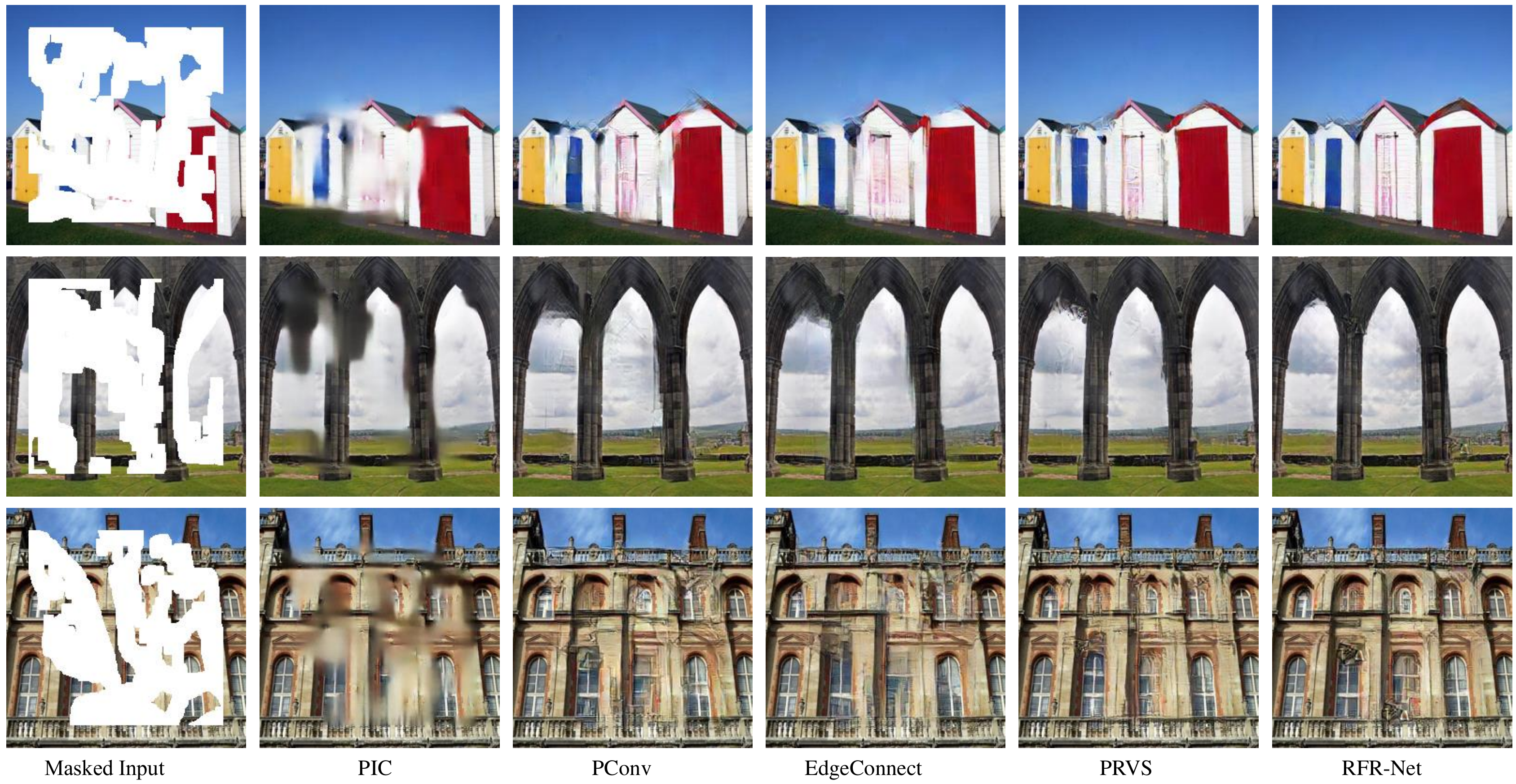}
\end{center}
\vspace{-6pt}
  \caption{Results on Places2}
\vspace{-12pt}
\label{fig:places}
\end{figure*}

\begin{figure}[t]
\begin{center}
   \includegraphics[width=1\linewidth]{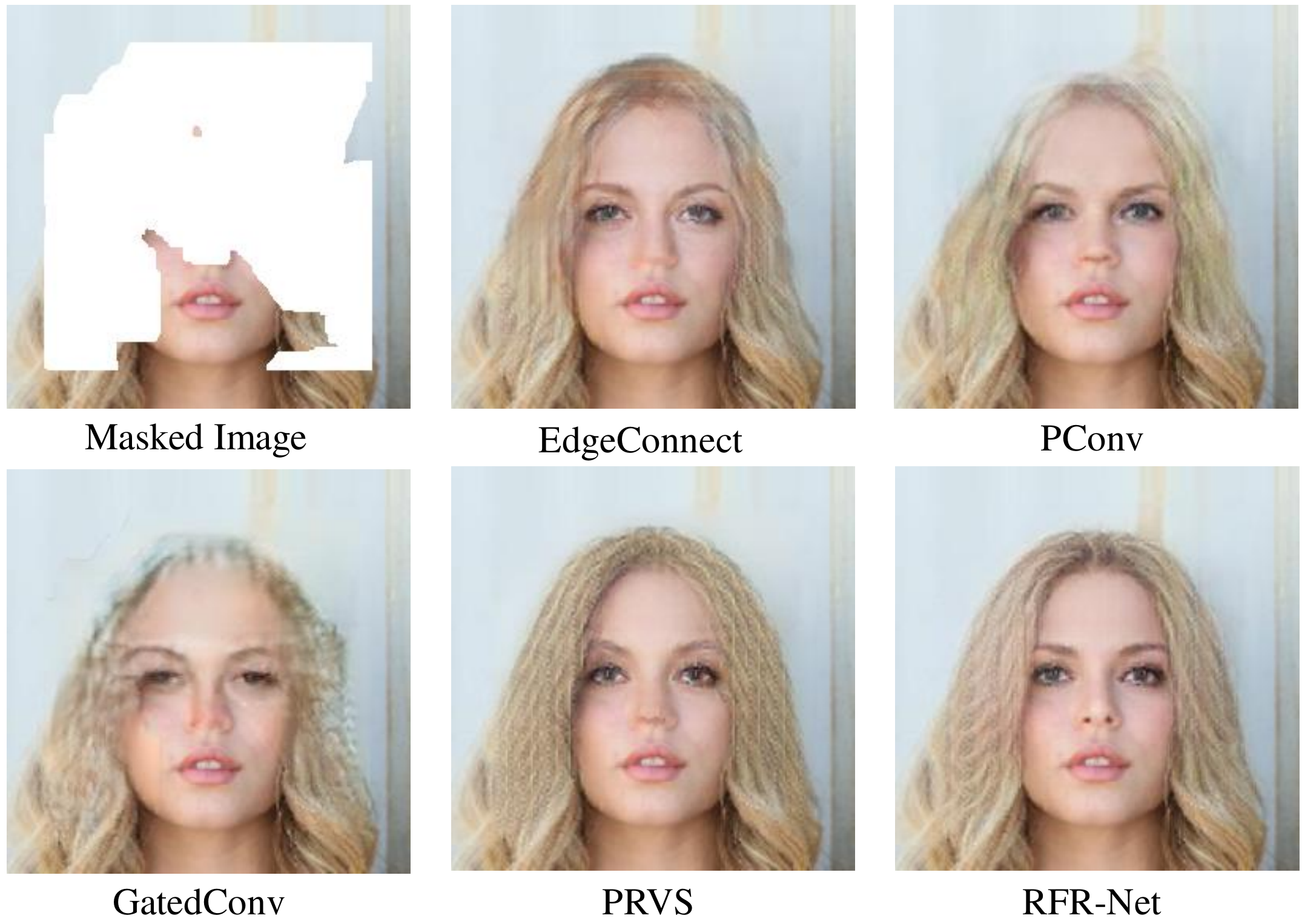}
\end{center}
\vspace{-8pt}
   \caption{Results on CelebA}
\vspace{-14pt}
\label{fig:celeba}
\end{figure}

\begin{figure}[t]
\begin{center}
   \includegraphics[width=1\linewidth]{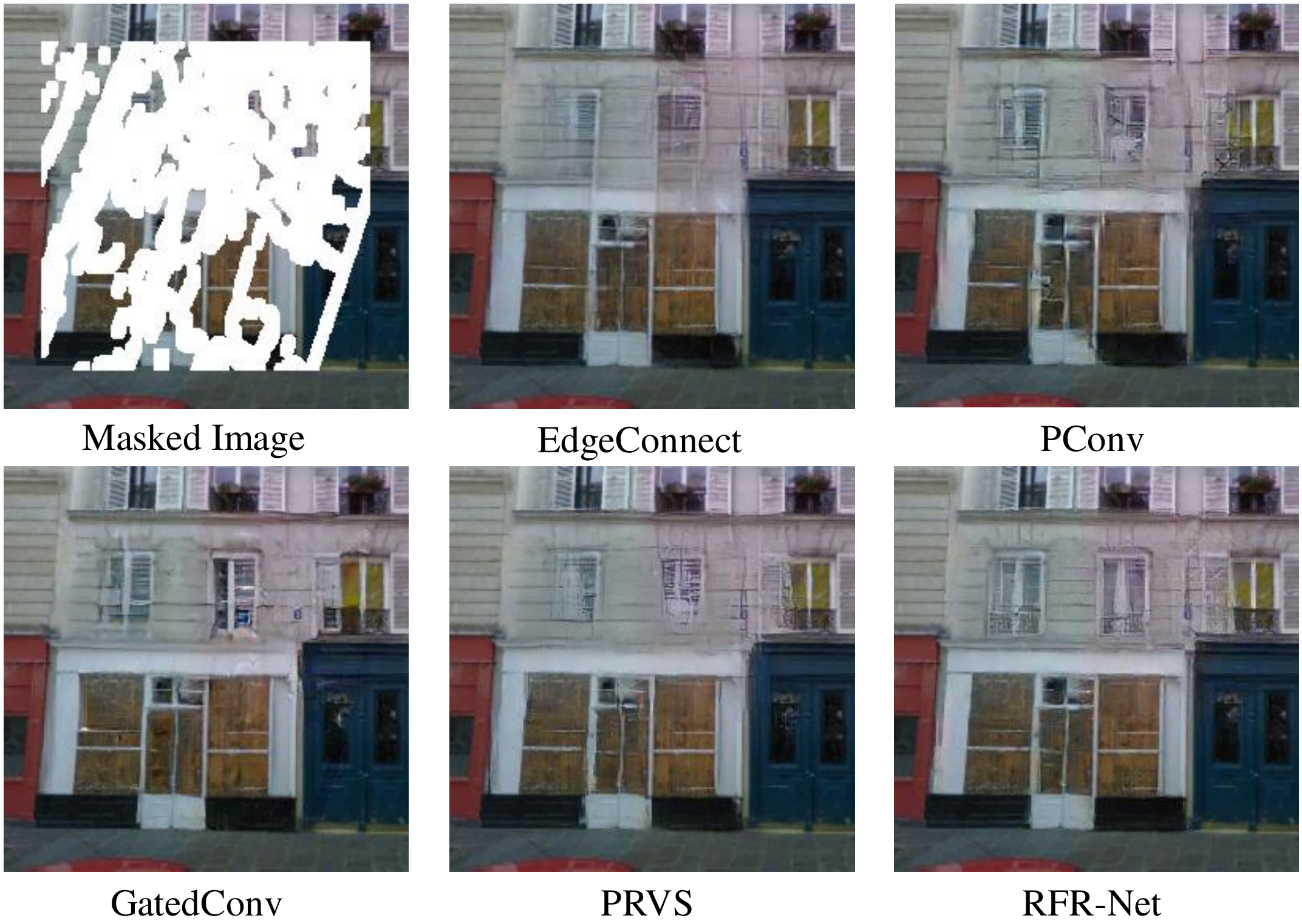}
\end{center}
\vspace{-8pt}
   \caption{Results on Paris StreetView}
\vspace{-14pt}
\label{fig:paris}
\end{figure}

\subsection{Training Setting}
We train our model with batch size 6 using the Adam optimizer. Since we only have a generator network to update, no optimizer is required for the discriminator. At the start, we use a learning rate of $1e^{-4}$ to train the model and then use $1e^{-5}$ for fine-tuning the model. During fine-tuning, we freeze all the batch normalization layers. For the hyper-parameters, we use 6 for $\lambda_{hole}$, 1 for $\lambda_{valid}$, 0.1 for $\lambda_{perceptual}$, and 180 for $\lambda_{style}$. All experiments are conducted using Python on an Ubuntu 17.10 system, with an i7-6800K 3.40GHz CPU and an 11G NVIDIA RTX2080Ti GPU.

\subsection{Datasets}
We use three public image datasets commonly used for image inpainting tasks and a mask dataset \cite{liu2018} to validate our model.

\noindent\textbf{Places2 Challenge Dataset}~\cite{zhou2018}: A dataset released by MIT containing over 8,000,000 images from over 365 scenes, which is very suitable for building inpainting models as it enables the model to learn the distribution from many natural scenes.

\noindent\textbf{CelebA Dataset}~\cite{liu2015}: A dataset focusing on human face images and containing over 180,000 training images. The model trained on this dataset can easily be transferred to face editing/completion tasks. 

\noindent\textbf{Paris StreetView Dataset}~\cite{doersch2012}: A dataset containing 14,900 training images and 100 test images collected from street views of Paris. This dataset mainly focuses on the buildings in the city. 

\subsection{Comparison Models}
We compare our approach with several state-of-the-art methods. These models are trained until convergence with the same experiment settings as ours. These models are: PIC~\cite{zheng2019}, PConv~\cite{liu2018}, GatedConv~\cite{yu2019}, EdgeConnect~\cite{nazeri2019}, and PRVS~\cite{li2019}.

\section{Results}

We conduct experiments on the three datasets and measure qualitative and quantitative results to compare our model with previous methods. We then compare the bare RFR-Net (without the attention module) with the existing backbone networks with respect to model size and quantitative performance to demonstrate the efficiency of our proposed method. Further, we conduct ablation studies to examine the design detail of our model.

\subsection{Comparisons with State-of-the-art Methods}
\label{sec:Comparisons-results}
In this part, we compared our RFR-Net with several state-of-the-art methods mentioned in the last section. We conducted qualitative analysis and quantitative analysis respectively to demonstrate the superiority of our method.

\noindent\textbf{Qualitative Comparisons}
Figs.~\ref{fig:places},~\ref{fig:celeba}, and~\ref{fig:paris} compare our method with five state-of-the-art approaches on the Places2, CelebA, and Paris StreetView datasets, respectively. Our inpainting results have significantly fewer noticeable inconsistencies than the state-of-the-art methods in most cases, especially for large holes. Compared to the other methods, our proposed algorithm generates more semantically plausible and elegant results.

\noindent\textbf{Quantitative Comparisons}
We also compare our model quantitatively in terms of structural similarity index (SSIM), peak signal-to-noise ratio (PSNR) and mean $\mathit{l}_1$ loss.
Table~\ref{tab:Quantitative_comparison in datases} lists the results with different ratios of irregular masks for the three datasets.
As shown in Table~\ref{tab:Quantitative_comparison in datases}, our method produces excellent results and the highest SSIM, PSNR and mean $\mathit{l}_1$ loss on the Places2, CelebA, and Paris StreetView datasets. The missing results in the table are due to the limitation in computational resources.

\begin{table*}\footnotesize
	\centering
	\begin{tabular}{l|l|ccc|ccc|ccc}
		\hline
		\multicolumn{2}{c|}{Dataset} & \multicolumn{3}{|c|}{Places2} & \multicolumn{3}{|c|}{CelebA} & \multicolumn{3}{|c}{Paris Street View}\\
		\hline
		\multicolumn{2}{c|}{Mask Ratio} & 10\%-20\% & 30\%-40\% & 50\%-60\% & 10\%-20\% & 30\%-40\% & 50\%-60\% & 10\%-20\% & 30\%-40\% & 50\%-60\%  \\
		\hline
		\hline
		\multirow{6}{*}{SSIM$^{\star}$} 
		&PIC         & 0.932 & 0.786 & 0.494 & 0.965 & 0.881 & 0.672 & 0.930 & 0.785 & 0.519\\
		&PConv         & 0.934 & 0.803 & 0.555 & 0.977 & 0.922 & 0.791 & 0.947 & 0.835 & 0.619\\
		&GatedConv      & ----- & ----- & ----- & 0.973 & 0.914 & 0.767 & 0.953 & 0.849 & 0.621\\
		&EdgeConnect& 0.933 & 0.802 & 0.553 & 0.975 & 0.915 & 0.759 & 0.950 & 0.849 & 0.646\\
		&PRVS          & 0.936 & 0.810 & 0.574 & 0.978 & 0.926 & 0.799 & 0.953 & 0.854 & 0.659\\
		&RFR-Net(Ours)                & \textbf{0.939} & \textbf{0.819} & \textbf{0.596} & \textbf{0.981} & \textbf{0.934} & \textbf{0.819} & \textbf{0.954} & \textbf{0.862} & \textbf{0.681}\\
		\hline
		\multirow{6}{*}{PSNR$^{\star}$} 
		&PIC         & 27.14 & 21.72 & 17.17 & 30.67 & 24.74 & 19.29 & 29.35 & 23.97 & 19.52\\
		&PConv         & 27.29 & 22.12 & 18.29 & 32.77 & 26.94 & 22.14 & 30.76 & 25.46 & 21.39\\
		&GatedConv      & ----- & ----- & ----- & 32.56 & 26.72 & 21.47 & 31.32 & 25.54 & 20.61\\
		&EdgeConnect& 27.17 & 22.18 & 18.35 & 32.48 & 26.62 & 21.49 & 31.19 & 26.04 & 21.89\\
		&PRVS           & 27.41 & 22.36 & 18.67 & 33.05 & 27.24 & 22.37 & 31.49 & 26.17 & 22.07\\
		&RFR-Net(Ours)                & \textbf{27.75} & \textbf{22.63} & \textbf{18.92} & \textbf{33.56} & \textbf{27.76} & \textbf{22.88} & \textbf{31.71} & \textbf{26.44} & \textbf{22.40}\\
		%&Mean $\mathit{l}_1^{\dag}$ & 0.1-0.2 & 0.3-0.4 & 0.5-0.6 & 0.1-0.2 & 0.3-0.4 & 0.5-0.6 & 0.1-0.2 & 0.3-0.4 & 0.5-0.6\\
		\hline
		\multirow{6}{*}{Mean $\mathit{l}_1^{\dag}$} 
		&PIC          & 0.0161 & 0.0441 & 0.0944 & 0.0111 & 0.0314 & 0.0749 & 0.0140 & 0.0379 & 0.0799\\
		&PConv          & 0.0154 & 0.0409 & 0.0824 & 0.0083 & 0.0236 & 0.0524 & 0.0123 & 0.0313 & 0.0623\\
		&GatedConv       & ------ & ------ & ------ & 0.0088 & 0.0245 & 0.0561 & 0.0120 & 0.0309 & 0.0660\\
		&EdgeConnect & 0.0157 & 0.0408 & 0.0821 & 0.0088 & 0.0247 & 0.0572 & 0.0110 & 0.0286 & 0.0582\\
		&PRVS            & 0.0148 & 0.0390 & 0.0778 & 0.0079 & 0.0224 & 0.0500 & 0.0111 & 0.0281 & 0.0562\\
		&RFR-Net(Ours)                 & \textbf{0.0142} & \textbf{0.0381} & \textbf{0.0761} & \textbf{0.0075} & \textbf{0.0212} & \textbf{0.0470} & \textbf{0.0110} & \textbf{0.0275} & \textbf{0.0546}\\
		\hline
	\end{tabular}
	\caption{Numerical comparison on three datasets. $^{\star}$Higher is better. $^{\dag}$Lower is better.}
	\vspace{-8pt}
	\label{tab:Quantitative_comparison in datases}
\end{table*}

\subsection{Model Efficiency}
As shown in Table~\ref{tab:backbone}, our bare RFR-Net (without the attention module) has fewer parameters than the widely used Coarse-To-Fine~\cite{liu2019} and PConv-UNet~\cite{liu2018,ronneberger2015} backbones. We compare the quantitative performance of these three networks (Table~\ref{tab:backbone}) with mask ratio of 40\%-50\% on Paris StreetView. For Coarse-To-Fine, we use the data reported in the CSA paper~\cite{liu2019} because the official code is unavailable. Our bare RFR-Net produces good results, with the best SSIM and PSNR of the tested methods. The inference time of our model for each image is usually between 85 and 95 ms, which is also faster than several state-of-the-art methods (e.g. \cite{yu2019, liu2019, li2019}). Taken together, we can conclude that our method can achieve better results than methods with similarly sized models, highlighting the efficiency of our proposed method.

\begin{table}\footnotesize
	\centering
	\begin{tabular}{l|cccc}
		\hline
		Backbone           & Coarse-To-Fine         & PConv           & bare RFR-Net\\
		\hline 
		Model Size         & 100M+                   & 33M             & 31M\\
		SSIM               & 0.768                  & 0.759           & 0.796 \\
		PSNR               & 23.10                  & 23.69           & 24.60\\
		\hline
	\end{tabular}
	\caption{Model Size of different backbones and their quantitative performances on Paris StreetView dataset of 40\%-50\% mask.}
	\label{tab:backbone}
\end{table}

\subsection{Ablation Studies}
In this section, we would like to verify the effect of our contributions separately. Here we mainly illustrate the effectiveness of KCA module and the influences of the recurrence number \textbf{\emph{IterNum}}. Due to the space limitation, more ablation studies are placed in the supplementary materials, including 1) moving the RFR module 2) the effect of feature merge and 3) applying the RFR module in other models.

\begin{figure}[t]
\begin{center}
   \includegraphics[width=1\linewidth]{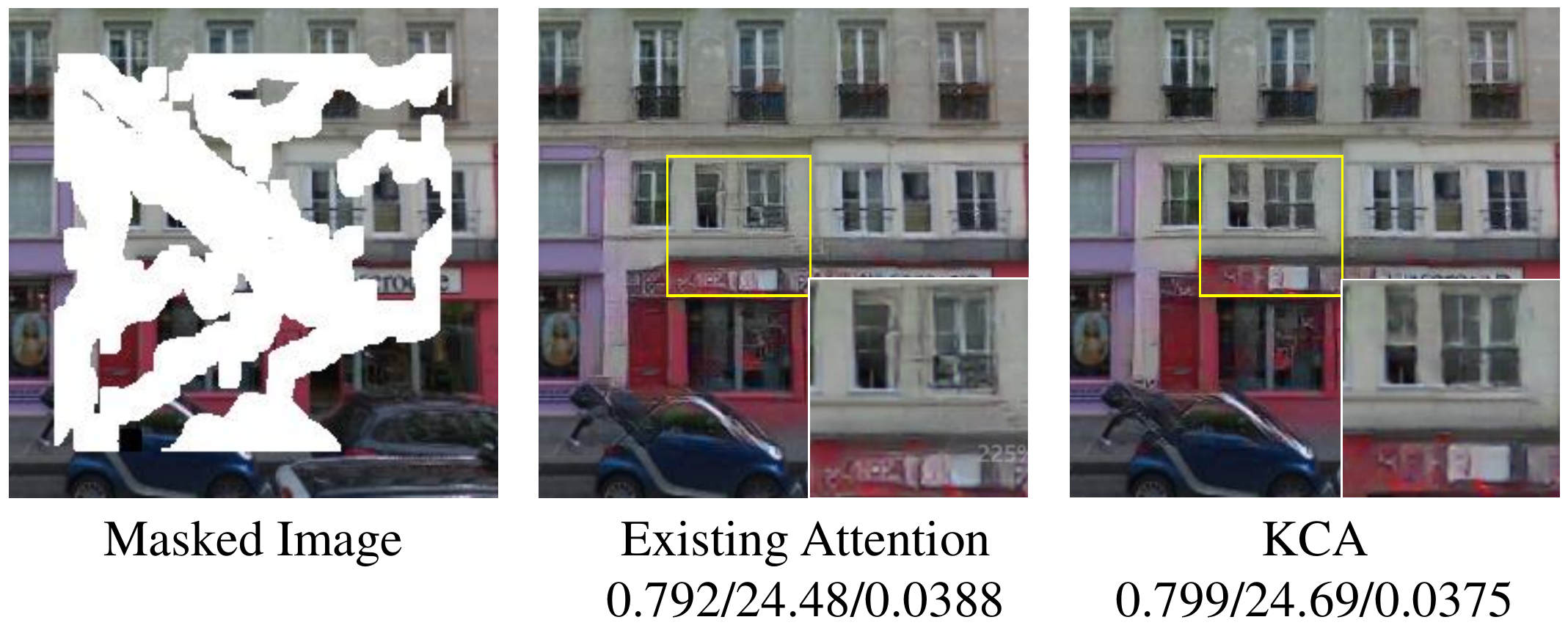}
\end{center}
\vspace{-6pt}
   \caption{Comparison results for different attention manners. From the left to the right are: (a) Input, (b) Existing Attention, (c) Knowledge Consistent Attention.}
\vspace{-8pt}
\label{fig:attention-manner}
\end{figure}

\noindent\textbf{Effect of Knowledge Consistent Attention}
As mentioned in Section~\ref{sec:ksa}, directly using the attention module from \cite{yu2018} in RFR module results in images with some boundary artifacts, as shown in Fig.~\ref{fig:attention-manner}. This is because the inconsistency of the feature map might lead to shadow-like artifacts after the feature merging process. The SSIM, PSNR, and Mean $\mathit{l}_1$ below these images demonstrate the quantitative comparisons between attention modules. These numbers are calculated from all images on Paris StreetView given the mask ratio of 40\%-50\%.

\noindent\textbf{The effect of recurrence number \textbf{\emph{IterNum}}} The results on Paris dataset corresponding to different \textbf{\emph{IterNum}}s after the same training iterations are given in Table. \ref{tab:iternum}. This ablation study reveals that our method is robust to the change of this hyper-parameter. The results also show that the improved performance compared to previous method is not from the deeper layers but from a more efficient architecture, since more \textbf{\emph{IterNum}}s do not improve the performance and our model has a smaller size than SOTAs.
\begin{table}[t]\footnotesize
	\centering
	\begin{tabular}{c|ccc}
		\hline
		\textbf{\emph{IterNum}} & 6 & 7 & 8\\
		\hline
		SSIM/PSNR & 0.857/26.26 & 0.852/26.07 & 0.854/26.15\\
		\hline
	\end{tabular}
	\caption{The influences of different \textbf{\emph{IterNum}}s, the number indicates the chosen number of recurrences.}
	\label{tab:iternum}		
	\vspace{-8pt}
\end{table}

\section{Conclusion}
In this paper, we propose the Recurrent Feature Reasoning network (RFR-Net) which progressively enriches information for the masked region and gives semantically explicit inpainted results. Further, a Knowledge Consistent Attention module is developed to assist the inference process of the RFR module. Extensive quantitative and qualitative comparisons, efficiency analysis, and ablation studies are conducted, which demonstrates the superiority of the proposed RFR-Net in both performance and efficiency.

\section{Acknowlgement}
This work was supported by the National Natural Science Foundation of China under grants 61822113 and 61771349, the Science and Technology Major Project of Hubei Province (Next-Generation AI Technologies) under Grant 2019AEA170, and Australian Research Council Project FL-170100117.

\clearpage
{\small
\bibliographystyle{ieee_fullname}
\bibliography{egbib}

\begin{thebibliography}{10}\itemsep=-1pt

\bibitem{barnes2009}
Connelly Barnes, Eli Shechtman, Adam Finkelstein, and Dan~B Goldman.
\newblock Patchmatch: A randomized correspondence algorithm for structural
  image editing.
\newblock {\em ACM TOG}, 28(3):24, 2009.

\bibitem{bertalmio2003}
Marcelo Bertalmio, Luminita Vese, Guillermo Sapiro, and Stanley Osher.
\newblock Simultaneous structure and texture image inpainting.
\newblock {\em IEEE TIP}, 12(8):882--889, 2003.

\bibitem{chan2001}
Tony~F Chan and Jianhong Shen.
\newblock Nontexture inpainting by curvature-driven diffusions.
\newblock {\em JVCIR}, 12(4):436--449, 2001.

\bibitem{criminisi2004}
Antonio Criminisi, Patrick P{\'e}rez, and Kentaro Toyama.
\newblock Region filling and object removal by exemplar-based image inpainting.
\newblock {\em IEEE TIP}, 13(9):1200--1212, 2004.

\bibitem{ding2019}
Ding Ding, Sundaresh Ram, and Jeffrey~J Rodr{\'\i}guez.
\newblock Image inpainting using nonlocal texture matching and nonlinear
  filtering.
\newblock {\em IEEE TIP}, 28(4):1705--1719, 2019.

\bibitem{doersch2012}
Carl Doersch, Saurabh Singh, Abhinav Gupta, Josef Sivic, and Alexei Efros.
\newblock What makes paris look like paris?
\newblock {\em ACM TOG}, 31(4):101, 2012.

\bibitem{fang2019}
Yixiang Fang, Kaiqiang Yu, Reynold Cheng, Laks~V.S. Lakshmanan, and Xuemin Lin.
\newblock Efficient algorithms for densest subgraph discovery.
\newblock In {\em Proc. VLDB}, pages 1719--1732, 2019.

\bibitem{goodfellow2014}
Ian~J. Goodfellow, Jean Pouget{-}Abadie, Mehdi Mirza, Bing Xu, David
  Warde{-}Farley, Sherjil Ozair, Aaron~C. Courville, and Yoshua Bengio.
\newblock Generative adversarial nets.
\newblock In {\em Proc. NIPS}, pages 2672--2680, 2014.

\bibitem{guo2019}
Zongyu Guo, Zhibo Chen, Tao Yu, Jiale Chen, and Sen Liu.
\newblock Progressive image inpainting with full-resolution residual network.
\newblock In {\em Proc. ACM MM}, pages 85--100, 2019.

\bibitem{iizuka2017}
Satoshi Iizuka, Edgar Simo-Serra, and Hiroshi Ishikawa.
\newblock Globally and locally consistent image completion.
\newblock {\em ACM TOG}, 36(4):107, 2017.

\bibitem{li2019}
Jingyuan Li, Fengxinag He, Lefei Zhang, Bo Du, and Dacheng Tao.
\newblock Progressive reconstruction of visual structure for image inpainting.
\newblock In {\em Proc. ICCV}, pages 6721--6729, 2019.

\bibitem{li2016}
Kangshun Li, Yunshan Wei, Zhen Yang, and Wenhua Wei.
\newblock Image inpainting algorithm based on {TV} model and evolutionary
  algorithm.
\newblock {\em Soft Comput.}, 20(3):885--893, 2016.

\bibitem{liu2018}
Guilin Liu, Fitsum~A Reda, Kevin~J Shih, Ting-Chun Wang, Andrew Tao, and Bryan
  Catanzaro.
\newblock Image inpainting for irregular holes using partial convolutions.
\newblock In {\em Proc. ECCV}, pages 85--100, 2018.

\bibitem{liu2019}
Hongyu Liu, Bin Jiang, Yi Xiao, and Chao Yang.
\newblock Coherent semantic attention for image inpainting.
\newblock In {\em Proc. ICCV}, pages 4170--4179, 2019.

\bibitem{liu2015}
Ziwei Liu, Ping Luo, Xiaogang Wang, and Xiaoou Tang.
\newblock Deep learning face attributes in the wild.
\newblock In {\em Proc. ICCV}, pages 3730--3738, 2015.

\bibitem{mirza2014}
Mehdi Mirza and Simon Osindero.
\newblock Conditional generative adversarial nets.
\newblock {\em CoRR}, abs/1411.1784, 2014.

\bibitem{nazeri2019}
Kamyar Nazeri, Eric Ng, Tony Joseph, Faisal Qureshi, and Mehran Ebrahimi.
\newblock Edgeconnect: Structure guided image inpainting using edge prediction.
\newblock In {\em Proc. ICCV Workshops}, 2019.

\bibitem{oh2019}
Seoung~Wug Oh, Sungho Lee, Joon{-}Young Lee, and Seon~Joo Kim.
\newblock Onion-peel networks for deep video completion.
\newblock In {\em Proc. ICCV}, pages 4403--4412, 2019.

\bibitem{pathak2016}
Deepak Pathak, Philipp Kr{\"{a}}henb{\"{u}}hl, Jeff Donahue, Trevor Darrell,
  and Alexei~A. Efros.
\newblock Context encoders: Feature learning by inpainting.
\newblock In {\em Proc. CVPR}, pages 2536--2544, 2016.

\bibitem{perez2003}
Patrick P{\'e}rez, Michel Gangnet, and Andrew Blake.
\newblock Poisson image editing.
\newblock In {\em Proc. SIGGRAPH}, pages 313--318, 2003.

\bibitem{ronneberger2015}
Olaf Ronneberger, Philipp Fischer, and Thomas Brox.
\newblock U-net: Convolutional networks for biomedical image segmentation.
\newblock In {\em Proc. MICCAI}, pages 234--241, 2015.

\bibitem{shetty2018}
Rakshith Shetty, Mario Fritz, and Bernt Schiele.
\newblock Adversarial scene editing: Automatic object removal from weak
  supervision.
\newblock In {\em Proc. NeurIPS}, pages 7717--7727, 2018.

\bibitem{song2018}
Linsen Song, Jie Cao, Linxiao Song, Yibo Hu, and Ran He.
\newblock Geometry-aware face completion and editing.
\newblock {\em CoRR}, 2018.

\bibitem{wang2019}
Ning Wang, Jingyuan Li, Lefei Zhang, and Bo Du.
\newblock Musical: Multi-scale image contextual attention learning for
  inpainting.
\newblock In {\em Proc. IJCAI}, pages 3748--3754, 2019.

\bibitem{xie2019}
Chaohao Xie, Shaohui Liu, Chao Li, Ming-Ming Cheng, Wangmeng Zuo, Xiao Liu,
  Shilei Wen, and Errui Ding.
\newblock Image inpainting with learnable bidirectional attention maps.
\newblock In {\em Proc. ICCV}, pages 8858--8867, 2019.

\bibitem{xiong2019}
Wei Xiong, Jiahui Yu, Zhe Lin, Jimei Yang, Xin Lu, Connelly Barnes, and Jiebo
  Luo.
\newblock Foreground-aware image inpainting.
\newblock In {\em Proc. CVPR}, 2019.

\bibitem{yan2018}
Zhaoyi Yan, Xiaoming Li, Mu Li, Wangmeng Zuo, and Shiguang Shan.
\newblock Shift-net: Image inpainting via deep feature rearrangement.
\newblock In {\em Proc. ECCV}, pages 3--19, 2018.

\bibitem{yeh2017}
Raymond~A Yeh, Chen Chen, Teck Yian~Lim, Alexander~G Schwing, Mark
  Hasegawa-Johnson, and Minh~N Do.
\newblock Semantic image inpainting with deep generative models.
\newblock In {\em Proc. CVPR}, pages 5485--5493, 2017.

\bibitem{yu2018}
Jiahui Yu, Zhe Lin, Jimei Yang, Xiaohui Shen, Xin Lu, and Thomas~S Huang.
\newblock Generative image inpainting with contextual attention.
\newblock In {\em Proc. CVPR}, pages 5505--5514, 2018.

\bibitem{yu2019}
Jiahui Yu, Zhe Lin, Jimei Yang, Xiaohui Shen, Xin Lu, and Thomas~S. Huang.
\newblock Free-form image inpainting with gated convolution.
\newblock In {\em Proc. ICCV}, pages 4471--4480, 2019.

\bibitem{zhang2018}
Haoran Zhang, Zhenzhen Hu, Changzhi Luo, Wangmeng Zuo, and Meng Wang.
\newblock Semantic image inpainting with progressive generative networks.
\newblock In {\em Proc. ACM MM}, pages 770--778, 2018.

\bibitem{zheng2019}
Chuanxia Zheng, Tat{-}Jen Cham, and Jianfei Cai.
\newblock Pluralistic image completion.
\newblock In {\em Proc. CVPR}, pages 1438--1447, 2019.

\bibitem{zhou2018}
Bolei Zhou, {\`{A}}gata Lapedriza, Aditya Khosla, Aude Oliva, and Antonio
  Torralba.
\newblock Places: A 10 million image database for scene recognition.
\newblock {\em IEEE TPAMI}, 40(6):1452--1464, 2018.

\end{thebibliography}
}

\clearpage
\onecolumn
\begin{center}
\LARGE \textbf{Supplementary Material}\\
%\vspace{0.3in}
\end{center}

\begin{appendix}
\vspace{-0.5cm}
\begin{table}[b]
\begin{tabular}{c|cccccccccc}
\hline
\multicolumn{9}{c}{RFR-Net Architecture}                                                                                        \\ \hline
Module Name   & Input\_Feat                 & In\_Size  & K\_Size & Stride  & Num\_Chan    & Out\_Size & BN        & Act\_Func  \\ \hline
PartialConv0  & Img\_Masked                 & Ori     & 7           & 2     & 64          & Ori/2     & T         & ReLU        \\
PartialConv1  & F\_Pconv0                   & Ori/2   & 7           & 1     & 64          & Ori/2     & T         & ReLU        \\
RFR Module    & F\_Pconv1                   & Ori/2   &             &       & 64          & Ori/2     & F         & None        \\
DeConv4       & F\_RFR                      & Ori/2   & 4           & 2     & 64          & Ori       & T         & Leaky\_ReLU \\ 
PartialConv4 & Cat(Img\_Masked, F\_Deconv4)& Ori     & 3           & 1     & 32          & Ori       & F         & Leaky\_ReLU \\ 
Conv9         & F\_Pconv3                   & Ori     & 3           & 1     & 32          & Ori       & T         & Leaky\_ReLU \\ 
Con10         & F\_Conv9                    & Ori     & 3           & 1     & 32          & Ori       & T         & Leaky\_ReLU \\ 
Output\_Conv  & Cat(F\_Pconv3, F\_Conv11)   & Ori     & 3           & 1     & 3           & Ori       & F         & None        \\ \hline\hline
\multicolumn{9}{c}{RFR Module Architecture}                                                                                     \\ \hline
Module Name   & Input\_Feat                 & In\_Size& K\_Size     & Stride& Num\_Chan   & Out\_Size & BN        & Act\_Func   \\ \hline
PartialConv2  & F\_Pconv1                   & Ori/2   & 7           & 1     & 64          & Ori/2     & F         & None        \\ 
PartialConv3  & F\_Pconv2                   & Ori/2   & 7           & 1     & 64          & Ori/2     & T         & ReLU        \\ 
Conv1         & F\_Pconv3                   & Ori/2   & 3           & 2     & 128         & Ori/4     & T         & ReLU        \\ 
Conv2         & F\_Conv1                    & Ori/4   & 3           & 2     & 256         & Ori/8     & T         & ReLU        \\ 
Conv3         & F\_Conv2                    & Ori/8   & 3           & 2     & 512         & Ori/16    & T         & ReLU        \\ 
Conv4         & F\_Conv3                    & Ori/16  & 3           & 1     & 512         & Ori/16    & T         & ReLU        \\
Conv5         & F\_Conv4                    & Ori/16  & 3           & 1     & 512         & Ori/16    & T         & ReLU        \\ 
Conv6         & F\_Conv5                    & Ori/16  & 3           & 1     & 512         & Ori/16    & T         & ReLU        \\ 
Conv7         & Cat(F\_Conv6, F\_Conv5)     & Ori/16  & 3           & 1     & 512         & Ori/16    & T         & Leaky\_ReLU \\ 
Conv8         & Cat(F\_Conv7, F\_Conv4)     & Ori/16  & 3           & 1     & 512         & Ori/16    & T         & Leaky\_ReLU \\ 
KCA           & F\_Conv8                    & Ori/16  &             &       & 512         & Ori/16    & F         & None     \\
DeConv1       & Cat(F\_KCA, F\_Conv3)       & Ori/16  & 4           & 2     & 256         & Ori/8     & T         & Leaky\_ReLU \\ 
DeConv2       & Cat(F\_Deconv1, F\_Conv2)   & Ori/8   & 4           & 2     & 128         & Ori/4     & T         & Leaky\_ReLU \\ 
DeConv3       & Cat(F\_Deconv2, F\_Conv1)   & Ori/4   & 4           & 2     & 64          & Ori/2     & T         & Leaky\_ReLU \\ 
Feature Merge & All F\_Deconv3              & Ori/2   &             &       & 64          & Ori/2     & F         & None        \\ \hline\hline
\end{tabular}
\caption{The architecture of the RFR-Net and RFR module respectively.}
\label{arch}
\end{table}
\section{Network Architecture}
The detailed architecture of our model is shown in Table \ref{arch}. The upper half of the table is the architecture of the whole RFR-Net. The bottom half is the architecture of the RFR Module. In specific, the meanings of the table are in the first row. Input\_Feat tells the source of the feature. In\_Size and Out\_Size indicate the sizes of the feature maps after they are processed and Ori means the size of the original input. K\_Size means the kernel size of the operators. Stride means the stride of the operators, which will be omitted if this parameter is not applicable. Num\_Chan means the channel number of the output feature map. BN means whether batch normalization layer is used after the operator. Act\_Fun means the non-linear function after the layer. Note that the RFR module actually has a recurrent design that feeds the feature map from layer "Deconv3" into "PartialConv1". Except for the first time recurrence, the input mask for "PartialConv1" comes from "PartialConv2" in the last recurrence. All leaky relu layers have a negative slope of 0.2.

\noindent\textbf{Connection to Partial Convolution U-Net:} The key component of the RFR-Net is the RFR module with a recurrent design that progressively recovers the damaged deep feature maps. To detect the area to be processed in next recurrence for RFR, there are several options, such as ``Partial Convolution", ``Gated Convolution" and ``Learnable Attention Maps". For simplicity, we used the layer and loss functions from Partial Convolution U-Net. However, the inpainting process of our RFR-Net differs largely from Partial Convolution U-Net. First, RFR-Net aims to generate high quality features in each recurrence so that the subsequent recurrences can benefit from them while the architecture from Partial Convolution U-Net propagates contents to the holes aggressively in each convolution layer, leading to information distortion during the process. Second, RFR-Net deploys shared parameters in each recurrence under a recurrent architecture while Partial Convolution U-Net calculates each step with different parameters under a feedforward architecture. Third, RFR-Net merges features generated from different recurrences to stabilize backward propagation by eliminating the gradient vanishing problem, while Partial Convolution U-Net does not have the gradient vanishing problem. Finally, we designed KCA to search for feature globally while Partial Convolution U-Net only captures information locally.

\noindent\textbf{The design of Feature Reasoning module:} The module aims to partly recover the masked region in deep feature maps. Therefore, we need to extract information from the already known features and estimate new contents. In this module, such new contents are produced in a feature reconstruction style with additional consideration. According to this objective, we develop our architecture based on the standard feature extraction and reconstruction technique, i.e., the encoder-and-decoder design.

\noindent\textbf{Computational Complexity:} 
The recurrent design of our RFR module increases the computational complexity compared to its non-recurrent version. However, since the RFR module is implemented for down-sampled feature maps (with much smaller size), the additional computation cost for each extra \textbf{\emph{IterNum}} during inference is very limited ($\sim$8ms and $\sim$20mb for each \textbf{\emph{IterNum}}).

\section{More Ablation Studies}
In this section of the supplementary material, we will show more ablation study about the RFR module. In specific, we first show that the RFR module can be installed in any part of an existing network and the computational cost can be controlled. Then we show that the RFR module can benefit a network whose input and output is not represented in the same space.
\subsection{Moving the RFR Module}
In this section, we will test how will the network's performance change if we move the RFR module up and down in the network. In specific, we will move the RFR module up and down in the network by adding or modifying the encoding and decoding layers to show that the RFR module can be flexibly installed into any part of a network. We tested four different models, which are using the RFR without downsampling layers (Modifying "PartialConv0" and "DeConv4" in the Table \ref{arch}), using the RFR after downsampling for once (original design in Table \ref{arch}), using the RFR after downsampling for twice (adding extra encoding and decoding layers before and after the RFR module respectively) and using the RFR after downsampling for three times. The RFR modules for each experiment remain the same to address any possible influence, which means the input feature maps all have 64 channels. Attention module is removed from all models we tested here. All results are from models trained on Paris StreetView dataset. For no-downsampling case, we are not able to train the network due to its unacceptably high computational cost and we only conduct memory cost experiment. By analyzing the Table \ref{movingrfr}, we notice that by moving the RFR module deeper, the computational cost is significantly reduced while the performances are only affected little. This further shows the superiority of the RFR module and the potential of the RFR-Net.
\begin{table}
	\centering
	\begin{tabular}{l|ccc|ccc|ccc|c}
		\hline
		\multicolumn{1}{c|}{Method} & \multicolumn{3}{|c}{SSIM$^{\star}$} & \multicolumn{3}{|c}{PSNR$^{\star}$} & \multicolumn{3}{|c}{Mean $l_1$$^{\dag}$} & \multicolumn{1}{|c}{Memory Usage} \\
		\hline
		\multicolumn{1}{c|}{Mask Ratio} & 0.1-0.2 & 0.3-0.4 & 0.5-0.6 & 0.1-0.2 & 0.3-0.4 & 0.5-0.6 & 0.1-0.2 & 0.3-0.4 & 0.5-0.6 & ANY  \\
		\hline
		DownSamp 0~           & \multicolumn{3}{|c}{Unable to train} &\multicolumn{3}{|c}{Unable to train}&\multicolumn{3}{|c|}{Unable to train}    & 1657M \\
		DownSamp 1~           & 0.955 & 0.860 & 0.673 & 31.72 & 26.38 & 22.29 & 0.0110 & 0.0279 & 0.0558 & 1122M\\
		DownSamp 2~           & 0.949 & 0.845 & 0.650 & 30.99 & 25.78 & 21.86 & 0.0118 & 0.0296 & 0.0582 & 863M\\
		DownSamp 3~           & 0.948 & 0.839 & 0.635 & 30.93 & 25.64 & 21.70 & 0.0129 & 0.0303 & 0.0600 & 834M\\
		\hline
	\end{tabular}
	\caption{Comparison between different places to put the module. $^{\star}$Higher is better. $^{\dag}$Lower is better.}
	\label{movingrfr}
\end{table}

\subsection{Effect of Feature Merging}
\begin{figure}[t]
\begin{center}
   \includegraphics[width=0.6\linewidth]{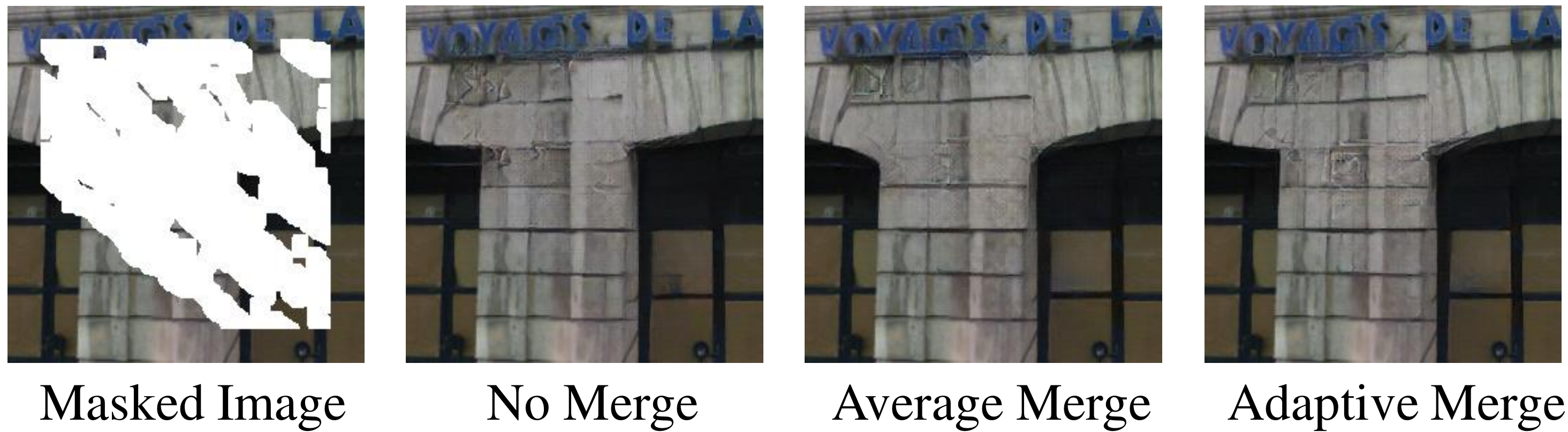}
\end{center}
\vspace{-6pt}
   \caption{Different methods for feature merging in RFR module. From the left to the right are: (a) Input, (b) No Merging, (c) Average Merging, (d) Adaptive Merging.}
\vspace{-8pt}
\label{fig:feature-merging}
\end{figure}
Fig.~\ref{fig:feature-merging} compares of different feature merging approaches in the RFR module on Paris StreetView. If only the last feature map is used as output (Fig.~\ref{fig:feature-merging} (b)) for feature merging, the texture is blurred and inadequate. This is because during the feature reasoning process, some feature generated in earlier recurrences might be damaged in order to further recover the hole region. Further, when we replace our adaptive merging (Fig.~\ref{fig:feature-merging} (d)) with average merging (Fig.~\ref{fig:feature-merging} (c)), the restored details of average merging are worse than those of adaptive merging. The reason is that when performing average merging, the feature values in the hole region are smoothed partly, as we explained in Sec. 3.1.3. This part demonstrates the advantage of the adaptive feature merging scheme we proposed.

\subsection{RFR Module For Structure Estimation}
\begin{figure}[b]
    \centering
    \subfigure[]{
    \begin{minipage}[b]{0.152\linewidth}
        \includegraphics[scale=0.31]{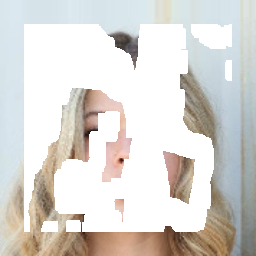}\vspace{2pt}
         \includegraphics[scale=0.31]{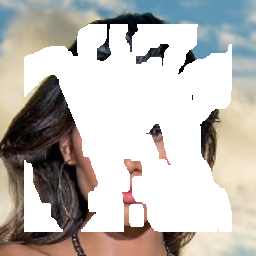}\vspace{2pt}
        \includegraphics[scale=0.31]{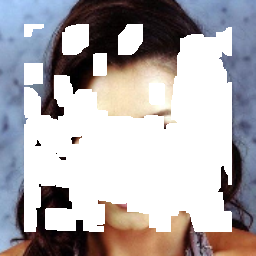}\vspace{2pt}
        \includegraphics[scale=0.31]{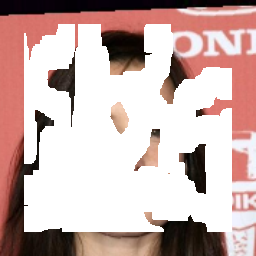}\vspace{2pt}
    \end{minipage}
    }
    \subfigure[]{
    \begin{minipage}[b]{0.152\linewidth}
        \includegraphics[scale=0.31]{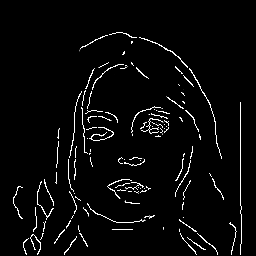}\vspace{2pt}
         \includegraphics[scale=0.31]{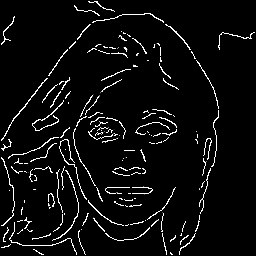}\vspace{2pt}
        \includegraphics[scale=0.31]{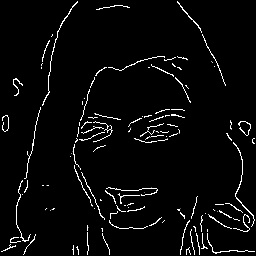}\vspace{2pt}
        \includegraphics[scale=0.31]{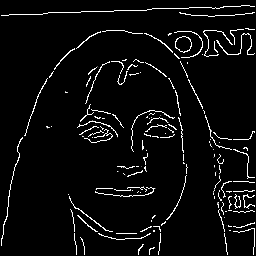}\vspace{2pt}
    \end{minipage}
    }
    \subfigure[]{
    \begin{minipage}[b]{0.152\linewidth}
        \includegraphics[scale=0.31]{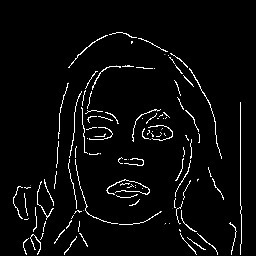}\vspace{2pt}
         \includegraphics[scale=0.31]{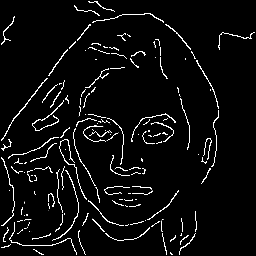}\vspace{2pt}
        \includegraphics[scale=0.31]{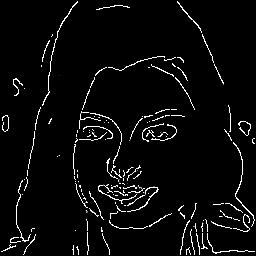}\vspace{2pt}
        \includegraphics[scale=0.31]{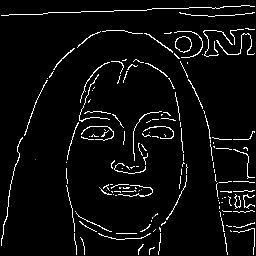}\vspace{2pt}
    \end{minipage}
    }
    \subfigure[]{
    \begin{minipage}[b]{0.152\linewidth}
         \includegraphics[scale=0.31]{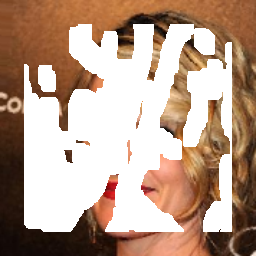}\vspace{2pt}
        \includegraphics[scale=0.31]{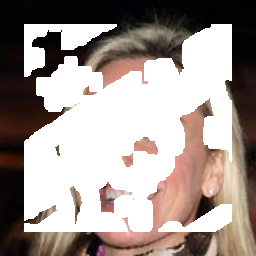}\vspace{2pt}
        \includegraphics[scale=0.31]{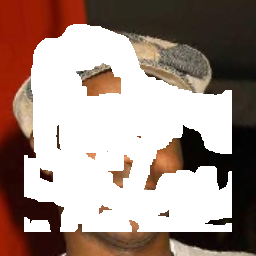}\vspace{2pt}
        \includegraphics[scale=0.31]{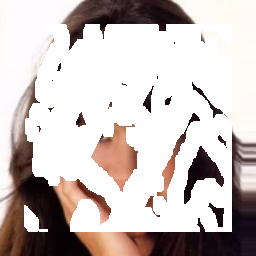}\vspace{2pt}
    \end{minipage}
    }
    \subfigure[]{
    \begin{minipage}[b]{0.152\linewidth}
         \includegraphics[scale=0.31]{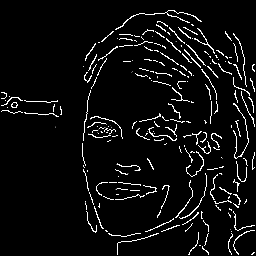}\vspace{2pt}
        \includegraphics[scale=0.31]{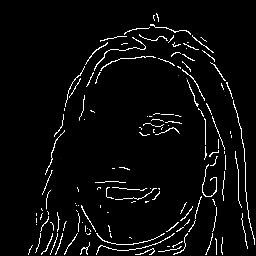}\vspace{2pt}
        \includegraphics[scale=0.31]{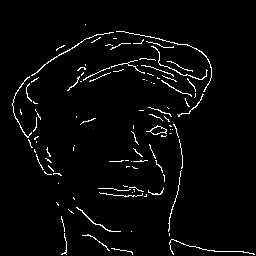}\vspace{2pt}
        \includegraphics[scale=0.31]{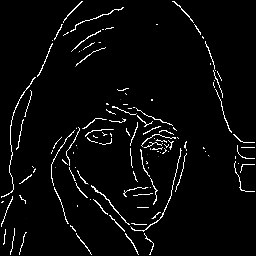}\vspace{2pt}
    \end{minipage}
    }
    \subfigure[]{
    \begin{minipage}[b]{0.152\linewidth}
         \includegraphics[scale=0.31]{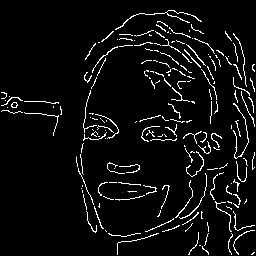}\vspace{2pt}
        \includegraphics[scale=0.31]{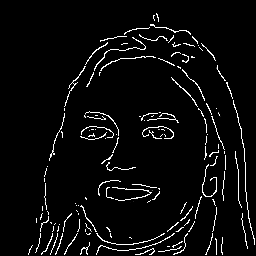}\vspace{2pt}
        \includegraphics[scale=0.31]{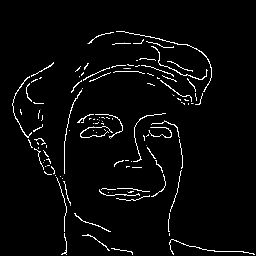}\vspace{2pt}
        \includegraphics[scale=0.31]{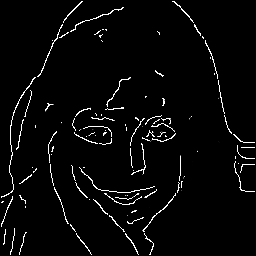}\vspace{2pt}
    \end{minipage}
    }
    \caption{RFR module for structure estimation.}
    \label{fig:edge}
\end{figure}
In this section, we will show that the RFR module can also boost the performance of a multi-stage method's subnetwork. In specific, the multi-stage method we choose is Edge-Connect, which first uses a network to reconstruct the boundary from the corrupted image and use the boundary to guide image inpainting. In this case, the each single network has different input and output space and therefore all existing progressive methods are not feasible. We replace the 8 residual blocks in the first network of the model with the RFR module and compare the results. The training settings are kept the same as that in the original paper of Edge-Connect. In Fig. \ref{fig:edge}, (a) and (d) are the masked input images. (b) and (e) are the results from Edge-Connect method's structure generator. (c) and (f) are the results from the RFR structure generator. The results from the RFR module are significantly better than the original edge generator. The results from this section also demonstrates the potential applications of the RFR net on other tasks where the input and output are not in the same representation space and some parts of the network include a encoder-decoder design architecture.

\clearpage
\section{More Results}
In this part, more visual comparisons and results are exhibited. In the first part, visual comparisons with state-of-the-art methods on CelebA and Paris StreetView datasets, which are omitted in the main text of the paper due to the space limitation. Then we show more visual results on three datasets with ground truth images. All these results demonstrates the effectiveness of our proposed methods.
\subsection{More Comparisons}
In this section, we show more comparison results. The results are on Paris StreetView and CelebA datasets. Our RFR-Net exhibits less boundary artifacts and much more explicit generated content.
\begin{figure}[h]
    \centering
    \subfigure[]{
    \begin{minipage}[b]{0.152\linewidth}
        \includegraphics[scale=0.31]{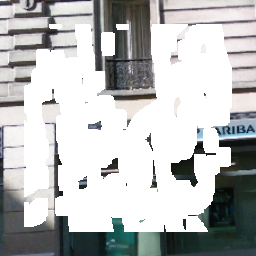}\vspace{2pt}
        \includegraphics[scale=0.31]{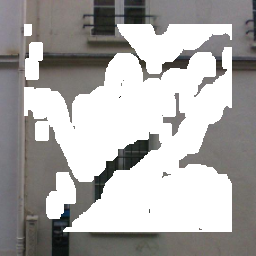}\vspace{2pt}
        \includegraphics[scale=0.31]{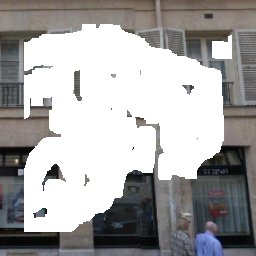}\vspace{2pt}
        \includegraphics[scale=0.31]{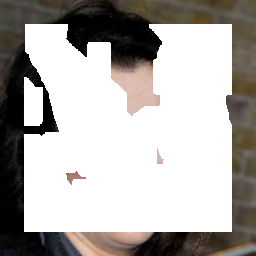}\vspace{2pt}
        \includegraphics[scale=0.62]{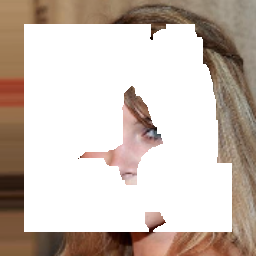}\vspace{2pt}
    \end{minipage}
    }
    \subfigure[]{
    \begin{minipage}[b]{0.152\linewidth}
        \includegraphics[scale=0.31]{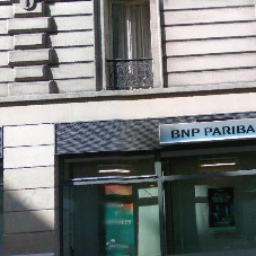}\vspace{2pt}
        \includegraphics[scale=0.31]{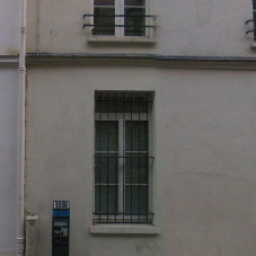}\vspace{2pt}
        \includegraphics[scale=0.31]{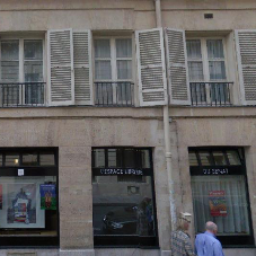}\vspace{2pt}
        \includegraphics[scale=0.31]{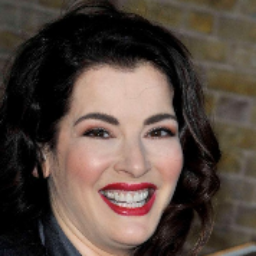}\vspace{2pt}
        \includegraphics[scale=0.31]{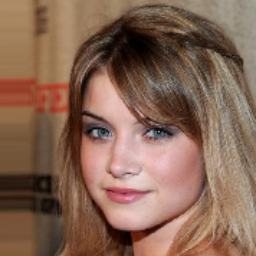}\vspace{2pt}
    \end{minipage}
    }    
    \subfigure[]{
        \begin{minipage}[b]{0.152\linewidth}
        \includegraphics[scale=0.31]{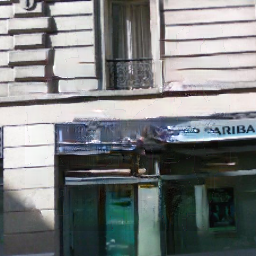}\vspace{2pt}
        \includegraphics[scale=0.31]{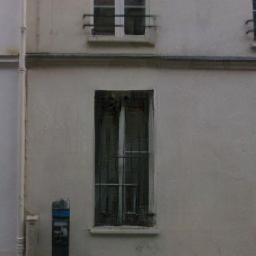}\vspace{2pt}
        \includegraphics[scale=0.31]{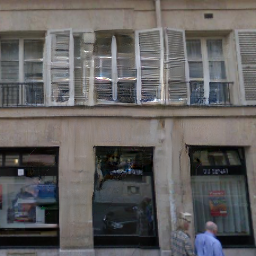}\vspace{2pt}
        \includegraphics[scale=0.31]{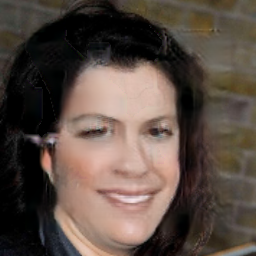}\vspace{2pt}
        \includegraphics[scale=0.31]{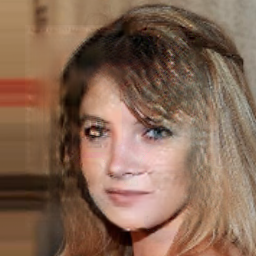}\vspace{2pt}
    \end{minipage}
    }
    \subfigure[]{
    \begin{minipage}[b]{0.152\linewidth}
        \includegraphics[scale=0.31]{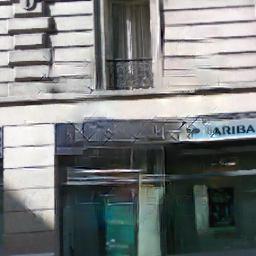}\vspace{2pt}
        \includegraphics[scale=0.31]{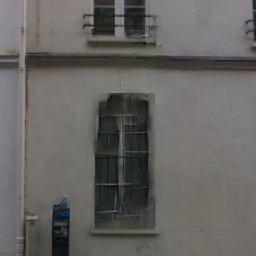}\vspace{2pt}
        \includegraphics[scale=0.31]{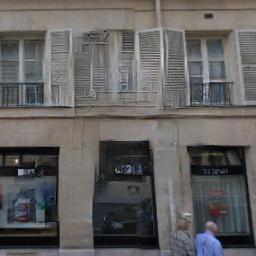}\vspace{2pt}
        \includegraphics[scale=0.31]{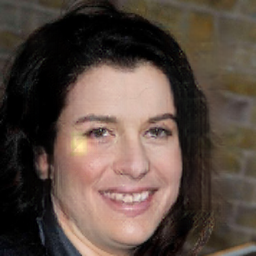}\vspace{2pt}
        \includegraphics[scale=0.62]{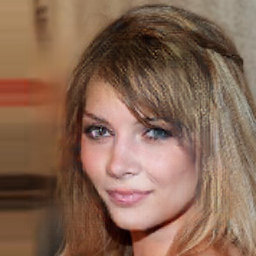}\vspace{2pt}
    \end{minipage}
    }
    \subfigure[]{
    \begin{minipage}[b]{0.152\linewidth}
        \includegraphics[scale=0.31]{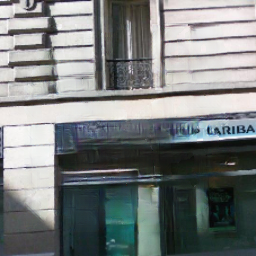}\vspace{2pt}
        \includegraphics[scale=0.31]{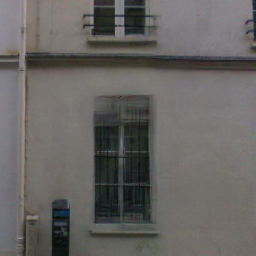}\vspace{2pt}
        \includegraphics[scale=0.31]{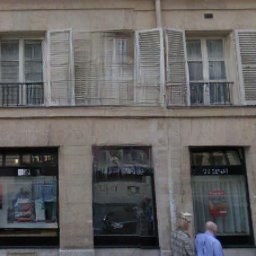}\vspace{2pt}
        \includegraphics[scale=0.31]{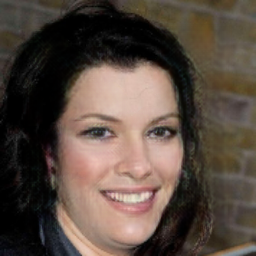}\vspace{2pt}
        \includegraphics[scale=0.62]{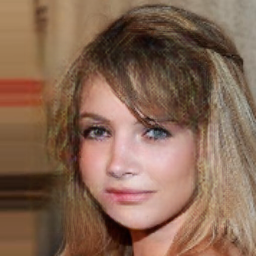}\vspace{2pt}
    \end{minipage}
    }
    \subfigure[]{
    \begin{minipage}[b]{0.152\linewidth}
        \includegraphics[scale=0.31]{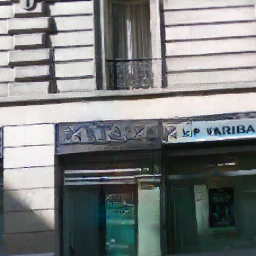}\vspace{2pt}
        \includegraphics[scale=0.31]{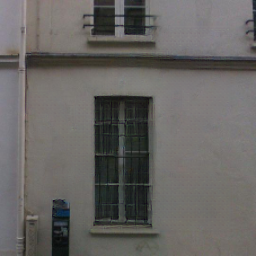}\vspace{2pt}
        \includegraphics[scale=0.31]{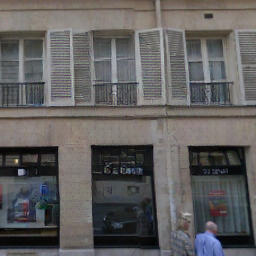}\vspace{2pt}
        \includegraphics[scale=0.31]{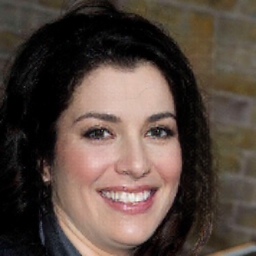}\vspace{2pt}
        \includegraphics[scale=0.31]{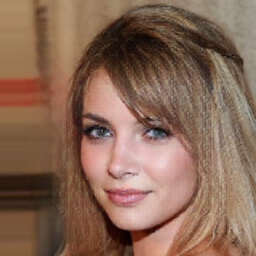}\vspace{2pt}
    \end{minipage}
    }
    \caption{More comparison results on Paris Street View and CelebA datasets. Our model stably produces well-structured results, even if the holes are large and challenging. From the left to the right are: (a) Input, (b) Ground Truth, (c) GatedConv, (d) PConv, (e) EdgeConnect and (f) Our RFR-Net.}
    \label{fig:places_results}
\end{figure}

\subsection{More Visual Results}
\begin{figure}[b]
    \centering
    \subfigure[]{
    \begin{minipage}[b]{0.152\linewidth}
        \includegraphics[scale=0.31]{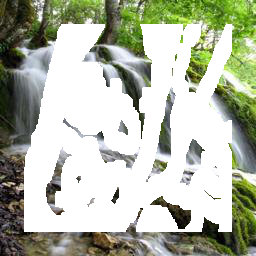}\vspace{2pt}
        \includegraphics[scale=0.31]{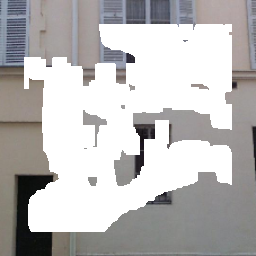}\vspace{2pt}
        \includegraphics[scale=0.31]{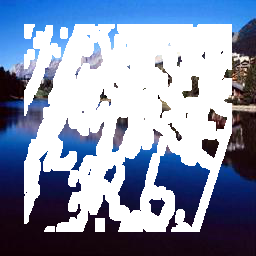}\vspace{2pt}
        \includegraphics[scale=0.517]{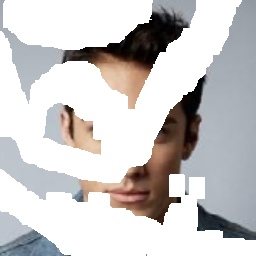}\vspace{2pt}
        \includegraphics[scale=0.31]{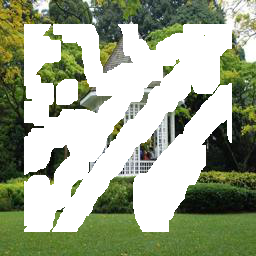}\vspace{2pt}

        \includegraphics[scale=0.31]{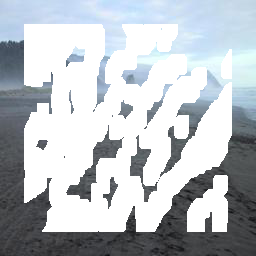}\vspace{2pt}
        \includegraphics[scale=0.31]{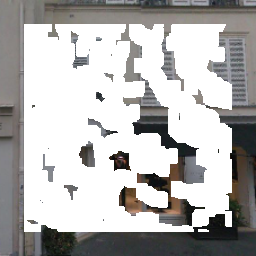}\vspace{2pt}
    \end{minipage}
    }
    \subfigure[]{
    \begin{minipage}[b]{0.152\linewidth}
        \includegraphics[scale=0.31]{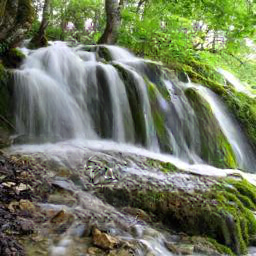}\vspace{2pt}
        \includegraphics[scale=0.31]{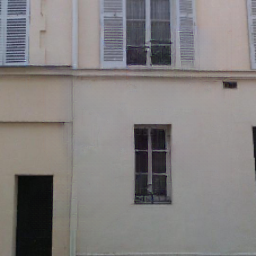}\vspace{2pt}
        \includegraphics[scale=0.31]{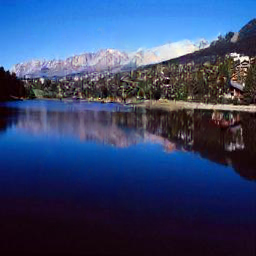}\vspace{2pt}
        \includegraphics[scale=0.31]{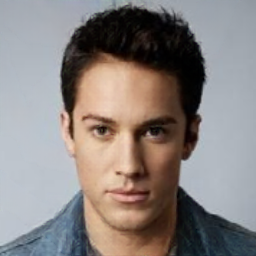}\vspace{2pt}
        \includegraphics[scale=0.31]{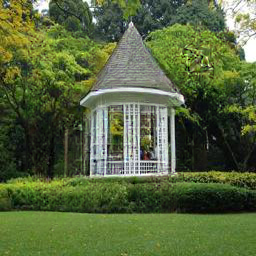}\vspace{2pt}

        \includegraphics[scale=0.31]{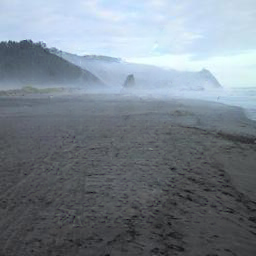}\vspace{2pt}
        \includegraphics[scale=0.31]{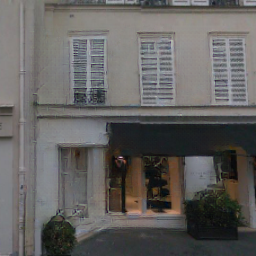}\vspace{2pt}
    \end{minipage}
    }
    \subfigure[]{
    \begin{minipage}[b]{0.152\linewidth}
        \includegraphics[scale=0.31]{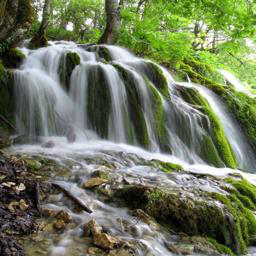}\vspace{2pt}
        \includegraphics[scale=0.31]{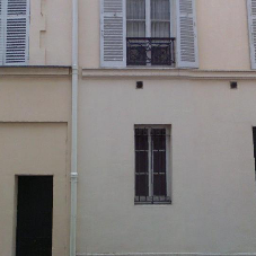}\vspace{2pt}
        \includegraphics[scale=0.31]{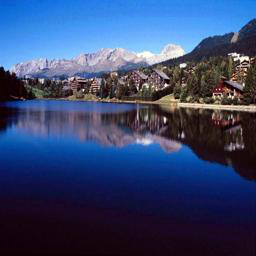}\vspace{2pt}
        \includegraphics[scale=0.31]{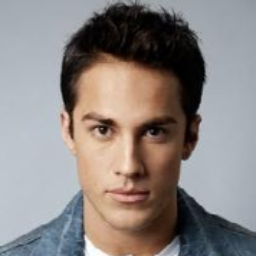}\vspace{2pt}
        \includegraphics[scale=0.31]{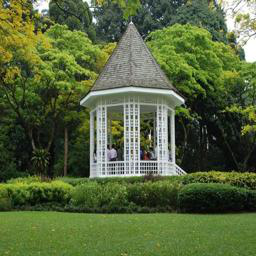}\vspace{2pt}

        \includegraphics[scale=0.31]{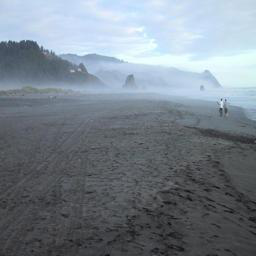}\vspace{2pt}
        \includegraphics[scale=0.31]{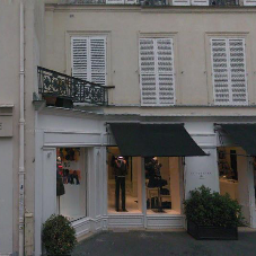}\vspace{2pt}
    \end{minipage}
    }
    \subfigure[]{
    \begin{minipage}[b]{0.152\linewidth}
        \includegraphics[scale=0.31]{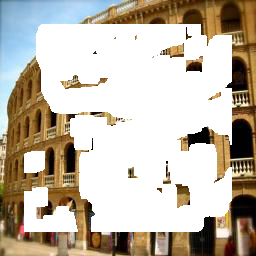}\vspace{2pt}
        \includegraphics[scale=0.31]{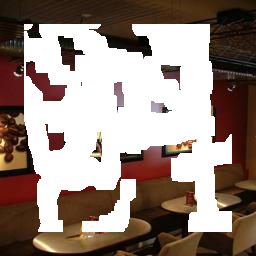}\vspace{2pt}
        \includegraphics[scale=0.31]{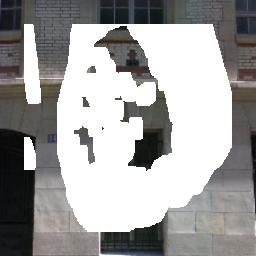}\vspace{2pt}
        \includegraphics[scale=0.31]{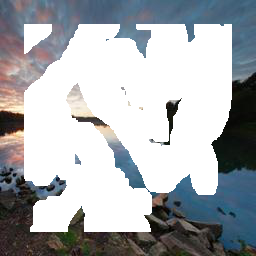}\vspace{2pt}
        \includegraphics[scale=0.31]{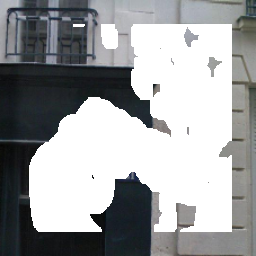}\vspace{2pt}
        \includegraphics[scale=0.31]{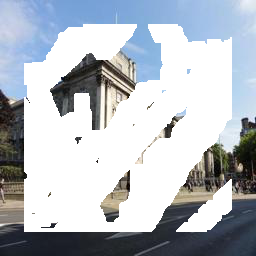}\vspace{2pt}
        \includegraphics[scale=0.517]{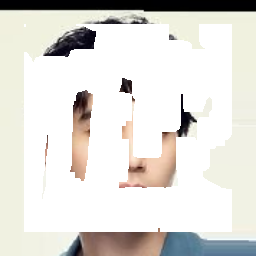}\vspace{2pt}

    \end{minipage}
    }
    \subfigure[]{
    \begin{minipage}[b]{0.152\linewidth}
        \includegraphics[scale=0.31]{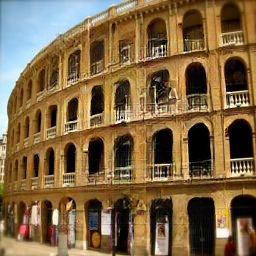}\vspace{2pt}
        \includegraphics[scale=0.31]{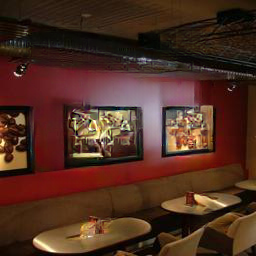}\vspace{2pt}
        \includegraphics[scale=0.31]{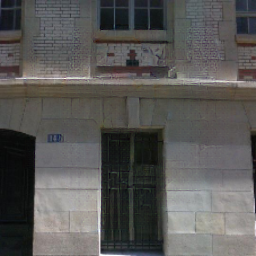}\vspace{2pt}
        \includegraphics[scale=0.31]{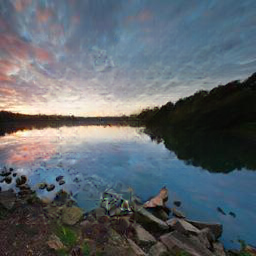}\vspace{2pt}
        \includegraphics[scale=0.31]{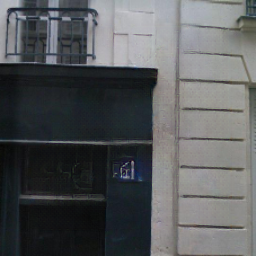}\vspace{2pt}
        \includegraphics[scale=0.31]{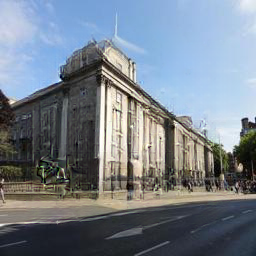}\vspace{2pt}
        \includegraphics[scale=0.31]{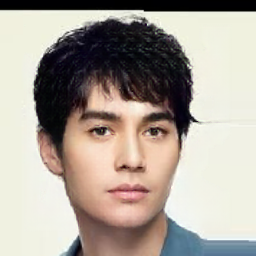}\vspace{2pt}

    \end{minipage}
    }
    \subfigure[]{
    \begin{minipage}[b]{0.152\linewidth}
        \includegraphics[scale=0.31]{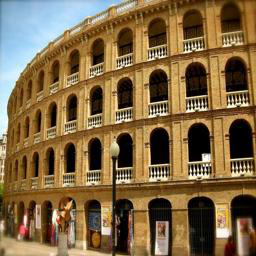}\vspace{2pt}
        \includegraphics[scale=0.31]{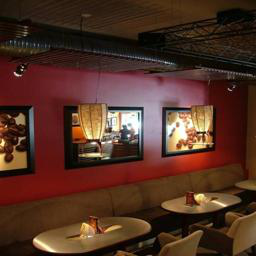}\vspace{2pt}
        \includegraphics[scale=0.31]{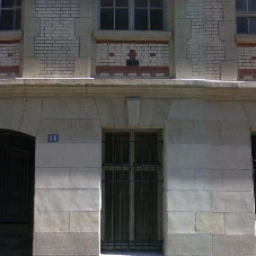}\vspace{2pt}
        \includegraphics[scale=0.31]{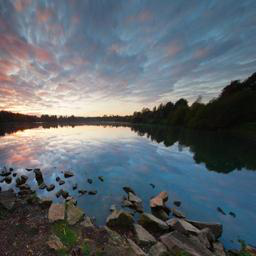}\vspace{2pt}
        \includegraphics[scale=0.31]{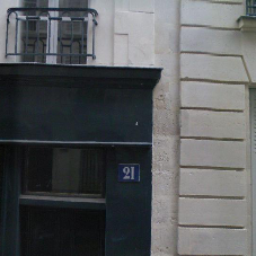}\vspace{2pt}
        \includegraphics[scale=0.31]{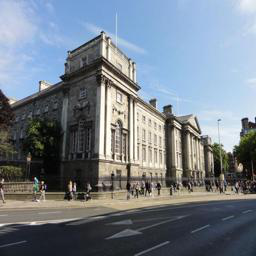}\vspace{2pt}
        \includegraphics[scale=0.31]{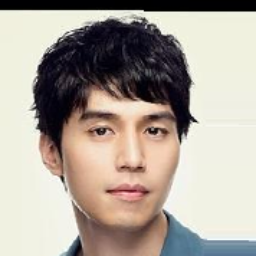}\vspace{2pt}

    \end{minipage}
    }
    \caption{More visual results. Our results have both coherent structures and plausible details.  From the left to the right are: (a) Input, (b) Our RFR-Net, (c) Ground Truth, (d) Input, (e) Our RFR-Net and (f) Ground Truth.}
    \label{fig:rfr_results}
\end{figure}

\end{appendix}

\end{document}